%% file: main.tex
\documentclass[10pt,twocolumn,letterpaper]{article}

\usepackage{iccv}              %
\usepackage{footnote}
\newcommand\blfootnote[1]{
  \begingroup
  \renewcommand\thefootnote{}\footnote{#1}
  \addtocounter{footnote}{-1}
  \endgroup
}

\input{preamble}

\newcommand{\basemodelname}{TiTok\xspace}
\newcommand{\modelname}{TA-TiTok\xspace}

\newcommand{\genmodelname}{MaskGen\xspace}

\definecolor{iccvblue}{rgb}{0.21,0.49,0.74}
\usepackage[pagebackref,breaklinks,colorlinks,allcolors=iccvblue]{hyperref}

\def\paperID{6497} %
\def\confName{ICCV}
\def\confYear{2025}

\title{Democratizing Text-to-Image Masked Generative Models with Compact Text-Aware One-Dimensional Tokens}

\author{
Dongwon Kim\textsuperscript{*$\dagger$1,2} \hspace{3mm}Ju He\textsuperscript{*1}  \hspace{3mm}Qihang Yu\textsuperscript{*1}  \\
\hspace{3mm}Chenglin Yang\textsuperscript{1} 
\hspace{3mm}Xiaohui Shen\textsuperscript{1}  \hspace{3mm}Suha Kwak\textsuperscript{2}  \hspace{3mm}Liang-Chieh Chen\textsuperscript{1}
\\
\textsuperscript{1} ByteDance Seed \hspace{8mm} \textsuperscript{2} POSTECH
\\
\small\url{https://tacju.github.io/projects/maskgen}
}

\begin{document}
\maketitle

\blfootnote{$^{\ast}$Equal contribution. $^\dagger$Work done as a research intern at ByteDance.}

\input{sec/0_abstract}    
\input{sec/1_intro}

\input{sec/2_related}

\input{sec/3_preliminary}

\input{sec/4_method}

\input{sec/5_experiment}

\input{sec/6_conclusion}

{
    \small
    \bibliographystyle{ieeenat_fullname}
    \bibliography{main}
}
\input{sec/X_suppl}

\end{document}

%% file: preamble.tex
\usepackage[utf8]{inputenc} %
\usepackage[T1]{fontenc}    %
\usepackage{url}            %
\usepackage{booktabs}       %
\usepackage{amsfonts}       %
\usepackage{nicefrac}       %
\usepackage{microtype}      %
\usepackage{xcolor}         %
\usepackage[accsupp]{axessibility}

\usepackage{bbding}
\usepackage{multirow}
\usepackage{algorithm}
\usepackage[noend]{algorithmic}
\usepackage{graphicx, amsmath, amssymb, caption, subcaption, overpic, textpos}
\usepackage{arydshln} %
\usepackage{listings}
\usepackage{setspace}

\usepackage{xspace}

\newcommand{\tabref}[1]{Tab.~\ref{#1}}

\def\eg{\emph{e.g}\onedot} 
\def\ie{\emph{i.e}\onedot} 
 
 \def\vs{\emph{vs}\onedot}
 
\def\etal{\emph{et al}\onedot}

\newcommand{\figref}[1]{Fig.~\ref{#1}}
\newcommand{\secref}[1]{Sec.~\ref{#1}}

\newlength\savewidth\newcommand\shline{\noalign{\global\savewidth\arrayrulewidth
  \global\arrayrulewidth 1pt}\hline\noalign{\global\arrayrulewidth\savewidth}}
\newcommand{\tablestyle}[2]{\setlength{\tabcolsep}{#1}\renewcommand{\arraystretch}{#2}\centering\footnotesize}

\usepackage{pifont}
\definecolor{mygreen}{RGB}{0,200,0}
\newcommand{\cmark}{\textcolor{mygreen}{\ding{51}}}%
\newcommand{\xmark}{\textcolor{red}{\ding{55}}}%

%% file: sec/0_abstract.tex
\begin{abstract}
Image tokenizers form the foundation of modern text-to-image generative models but are notoriously difficult to train. Furthermore, most existing text-to-image models rely on large-scale, high-quality private datasets, making them challenging to replicate. In this work, we introduce \textbf{T}ext-\textbf{A}ware \textbf{T}ransformer-based 1-D\textbf{i}mensional \textbf{Tok}enizer (\modelname), an efficient and powerful image tokenizer that can utilize either discrete or continuous 1-dimensional tokens. \modelname uniquely integrates textual information during the tokenizer decoding stage (\ie, de-tokenization), accelerating convergence and enhancing performance.
\modelname also benefits from a simplified, yet effective, one-stage training process, eliminating the need for the complex two-stage distillation used in previous 1-dimensional tokenizers. This design allows for seamless scalability to large datasets. Building on this, we introduce a family of text-to-image \textbf{Mask}ed \textbf{Gen}erative Models (\genmodelname), trained exclusively on open data while achieving comparable performance to models trained on private data.
We aim to release both the efficient, strong \modelname tokenizers and the open-data, open-weight \genmodelname models to promote broader access and democratize the field of text-to-image masked generative models.
\end{abstract}

%% file: sec/1_intro.tex
\section{Introduction}
\label{sec:intro}

\begin{figure*}[!ht]
    \centering
    \includegraphics[width=0.98\linewidth]{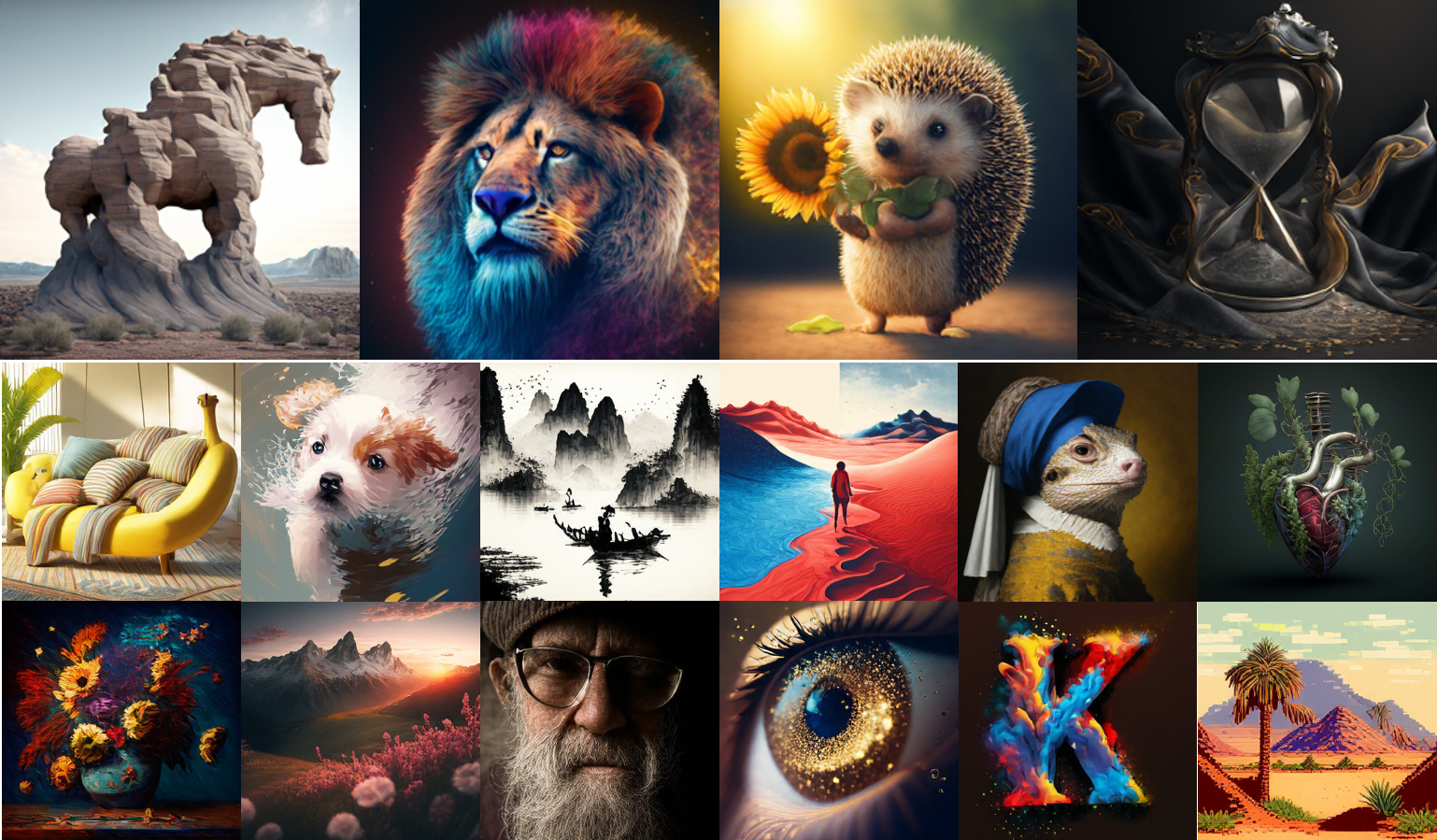}
    \caption{
    \textbf{Text-to-Image (T2I) Generation Results by \genmodelname.}
    \genmodelname, powered by the proposed compact text-aware 1D tokenizer \modelname, is an efficient masked generative model that achieves state-of-the-art performance on multiple T2I benchmarks using only open data. The open-data, open-weight \genmodelname models are designed to promote broader access and democratize T2I masked generative models.  
    }
    \vspace{-4mm}
    \label{fig:teaser}
\end{figure*}

In recent years, text-to-image generation has seen remarkable progress across various frameworks, including diffusion models~\cite{rombach2022high,saharia2022photorealistic,esser2024scaling,shin2025deeply,ren2025grouping,he2025flowtok,yang20241}, autoregressive visual models~\cite{gafni2022make,team2024chameleon,sun2024autoregressive,ren2025flowar,ren2025beyond}, and masked generative models~\cite{chang2023muse,bai2024meissonic,fan2024fluid}. A crucial component of these models is a robust image tokenizer—either discrete~\cite{van2017neural} or continuous~\cite{kingma2013auto}—which transforms images into tokenized representations. These models then incorporate text conditions with the tokenized representations using methods such as cross-attention~\cite{rombach2022high}, concatenation~\cite{bao2023all}, or conditioning embeddings~\cite{peebles2023scalable}, ultimately leveraging the tokenizer to generate high-quality images aligned with input text prompts.

Despite these advancements, replicating these models remains challenging due to the substantial resources required. Model sizes have grown from millions to billions of parameters, leading to prohibitively high training and inference costs. Furthermore, many of these models rely on large-scale, high-quality \textit{proprietary} image-text datasets, which are critical for achieving high-fidelity generation but prevent the research community from fully reproducing results. Given these obstacles, a pivotal question arises: \textit{Can we develop a text-to-image generative model that is both efficient and effective using only open data, enabling reproducibility?}

In this work, we address this question by building on the recent concept of compact one-dimensional tokenizers to facilitate both efficient and high-quality image tokenization. Traditional tokenizers~\cite{kingma2013auto,van2017neural,esser2021taming} rely on 2D grid-based latent representations, which struggle to handle the inherent redundancies in images, as neighboring patches often display similarities. Recently, Yu~\etal~\cite{yu2024image} introduced the \textbf{T}ransformer-based 1-D\textbf{i}mensional \textbf{Tok}enizer (TiTok), which efficiently tokenizes images into compact 1D latent sequences by removing the fixed correspondence between latent representation and 2D image patches (\ie, each 1D token can represent any region in an image, rather than being tied to a specific patch). This approach results in a highly efficient tokenizer, significantly improving sampling speed compared to previous methods~\cite{van2017neural,esser2021taming,chang2022maskgit,peebles2023scalable}.

However, extending TiTok to support text-to-image generation presents three main challenges: (1) its reliance on a complex two-stage training pipeline, which limits scalability to larger datasets necessary for diverse text-to-image generation beyond ImageNet~\cite{deng2009imagenet}; (2) its restriction to a Vector-Quantized (VQ) variant, leaving unexplored the potential benefits of a continuous Variational Autoencoder (VAE) representation; and (3) its focus on reconstructing low-level image details, which may lack the high-level semantics needed for effective alignment with textual descriptions.

To address these limitations, we introduce several key innovations. First, we streamline the training process for the 1D tokenizer by developing an efficient one-stage training procedure, removing the need for the complex two-stage pipeline used in the original framework~\cite{yu2024image}. This improvement enables scalable training of the 1D tokenizer on large-scale text-image datasets without multi-stage complexity.

Second, we extend the 1D tokens to continuous VAE representations, which allow for more consistent and accurate image reconstructions than the VQ counterpart. This approach combines the sampling efficiency of 1D tokens (due to the reduced number of tokens) with the improved reconstruction quality afforded by the continuous VAE representation, eliminating the quantization loss seen in VQ.

Third, we incorporate textual information during the de-tokenization stage to enhance semantic alignment with text prompts. Specifically, by concatenating CLIP~\cite{radford2021learning} embeddings of captions with the tokenized image representations, we enable higher-quality image reconstructions that better retain fine details. This approach powers \modelname, our novel and efficient \textbf{T}ext-\textbf{A}ware \textbf{T}ransformer-based 1-D\textbf{i}mensional \textbf{Tok}enizer, trained on the large-scale dataset (\eg, DataComp~\cite{gadre2024datacomp}) to capture a broad and diverse range of concepts.

Building upon \modelname, we introduce \genmodelname, a family of text-to-image \textbf{Mask}ed \textbf{Gen}erative models. \genmodelname is a versatile framework that supports both discrete and continuous token representations. For images represented by discrete tokens, \genmodelname is trained using cross-entropy loss~\cite{chang2022maskgit}, while for images with continuous tokens, it leverages the recent diffusion loss~\cite{li2024autoregressive}.
To encode captions as text conditioning, we utilize CLIP~\cite{radford2021learning}, instead of the more resource-intensive T5-XXL encoder~\cite{raffel2020exploring}, used in other recent text-to-image models~\cite{saharia2022photorealistic,chang2023muse,esser2024scaling}. Although effective, T5-XXL incurs significantly higher computational and storage costs. CLIP provides a more efficient alternative, making our approach more accessible to research groups with limited compute resources.
In terms of architecture, we adopt a straightforward design: we concatenate text conditions with image tokens before feeding them to the Diffusion Transformer~\cite{peebles2023scalable}, applying separate adaptive LayerNorm (adaLN~\cite{ba2016layer,peebles2023scalable}) parameters for text and image modalities, as in MM-DiT~\cite{esser2024scaling}. Additionally, we find it beneficial to incorporate the aesthetic score as an additional conditioning signal via adaLN. This allows for more nuanced control over the generated images, beyond text input alone.

To ensure reproducibility, we exclusively use images from publicly available sources, including DataComp~\cite{gadre2024datacomp}, LAION~\cite{schuhmann2022laion}, and CC12M~\cite{changpinyo2021conceptual}, as well as synthetic data from JourneyDB~\cite{sun2024journeydb} and DALL-E 3~\cite{betker2023improving,Egan_Dalle3_1_Million_2024}. Given the noisiness of web-sourced text-image pairs, we filter images based on aesthetic scores ($\ge$ 5), resolution (aspect ratio < 2 and longer side $\ge$ 256), and remove any images containing watermarks. Following the approach of DALL-E 3~\cite{betker2023improving}, we further enhance text quality by recaptioning high-aesthetic subsets from DataComp and subsets from LAION (LAION-pop and LAION-art~\cite{schuhmann2022laion}) using the state-of-the-art vision language model Molmo~\cite{deitke2024molmo}.

Notably, despite being trained entirely on publicly available datasets, \genmodelname achieves strong performance and efficiency in text-to-image generation.
On MJHQ-30K~\cite{playground-v2}, the lighter \genmodelname-L (568M) with discrete tokens achieves a generation FID of 7.74, outperforming Show-o~\cite{xie2024show} (14.99) with 30.3× faster inference throughput. It also surpasses SD-2.1~\cite{rombach2022high} (26.96) and PixArt-$\alpha$~\cite{chen2024pixartalpha} (9.85) while requiring only 2\% and 21\% of their training times, respectively.
Furthermore, the larger \genmodelname-XL (1.1B) achieves FID scores of 7.51 and 6.53 on MJHQ-30K~\cite{playground-v2} and overall scores of 0.57 and 0.55 on the GenEval~\cite{ghosh2024geneval} benchmark using discrete and continuous tokens, respectively.

To promote further research on text-to-image masked generative models, we will release the training code and model weights for both \modelname and \genmodelname. To our knowledge, \genmodelname is the first open-weight, open-data \textit{masked generative model} for text-to-image synthesis to achieve performance comparable to state-of-the-art models, advancing the democratized access to high-performance masked generative models in this field.

%% file: sec/2_related.tex
\section{Related Work}
\label{sec:related}

\noindent \textbf{Image Tokenization.}
Modern generative image models rely on image tokenization for efficient generation~\cite{esser2021taming,rombach2022high,chang2022maskgit,yu2022scaling}. During training, images are encoded into discrete~\cite{van2017neural} or continuous~\cite{kingma2013auto} tokens, allowing the model to focus on learning semantically meaningful information~\cite{rombach2022high} rather than directly working with pixels~\cite{gregor2014deep}.

Image tokenization approaches fall into two main paradigms. The first, discrete tokenization~\cite{van2017neural}, maps each token to a codebook entry and is well-suited to autoregressive~\cite{esser2021taming} or masked generative models~\cite{chang2022maskgit}, as it enables techniques directly from language models~\cite{gpt3}. Introduced in VQ-VAE~\cite{van2017neural} and later improved in VQ-GAN~\cite{esser2021taming} with adversarial loss~\cite{goodfellow2014generative}, this approach has been further scaled through advanced codebook management techniques~\cite{yu2021vector,yu2023language,zheng2023online}.

The second paradigm, continuous tokenization, follows the VAE~\cite{kingma2013auto} framework, enabling latent representations drawn from a normal distribution. While less common with masked generative models due to the simpler loss definitions with discrete tokens, continuous tokenization was recently adapted for such models in MAR~\cite{li2024autoregressive}, using a diffusion module to sample tokens from the normal distribution.

\noindent \textbf{Image Generation with Sequence Models.}
Initially developed for language tasks, sequence models like BERT~\cite{devlin2018bert} and GPT~\cite{gpt3} have been effectively adapted for image generation. Early approaches focused on autoregressive pixel generation~\cite{gregor2014deep,van2016conditional,parmar2018image,chen2020generative}, but recent methods model the joint distribution of image tokens, leading to two main approaches: autoregressive~\cite{esser2021taming} and masked generative models~\cite{chang2022maskgit}.

Autoregressive models predict tokens sequentially, following GPT’s strategy~\cite{ramesh2021zero,yu2022scaling,sun2024autoregressive,luo2024open,yu2024randomized}, while masked generative models adopt a BERT-like objective, predicting masked tokens simultaneously. This approach gives masked models a substantial edge in sampling speed, as they do not require token-by-token generation~\cite{chang2022maskgit,li2023mage,yu2023magvit,yu2023language,weber2024maskbit}.
Building on these efficiency benefits, our work develops an open-source masked generative model that leverages 1D tokenization for efficient text-to-image generation.

\noindent \textbf{Text-to-Image Generation.}
While diffusion models dominate text-to-image generation~\cite{rombach2022high,saharia2022photorealistic,bao2023all,peebles2023scalable,podell2023sdxl,chen2024pixartalpha,liu2024alleviating}, sequence models have shown strong potential as well~\cite{yu2022scaling,gafni2022make,chang2023muse,fan2024fluid}. Examples like Muse~\cite{chang2023muse} and Parti~\cite{yu2022scaling} demonstrate the success of masked generative and autoregressive approaches for generating high-quality images from text.
Recent innovations in diffusion models, such as improved architectures~\cite{peebles2023scalable,esser2024scaling}, micro-conditioning for finer control~\cite{podell2023sdxl}, and advanced image recaptioning for better text-image alignment~\cite{betker2023improving}, offer potential benefits for masked generative models. In our work, we integrate these recent improvements from diffusion models into the masked generative model framework. By incorporating these improvements, we aim to enhance the quality of generation while maintaining their inherent advantages in sampling efficiency.

%% file: sec/3_preliminary.tex
\section{Preliminary}
\label{sec:preliminary}

\begin{figure*}[!t]
    \centering
    \includegraphics[width=0.98\linewidth]{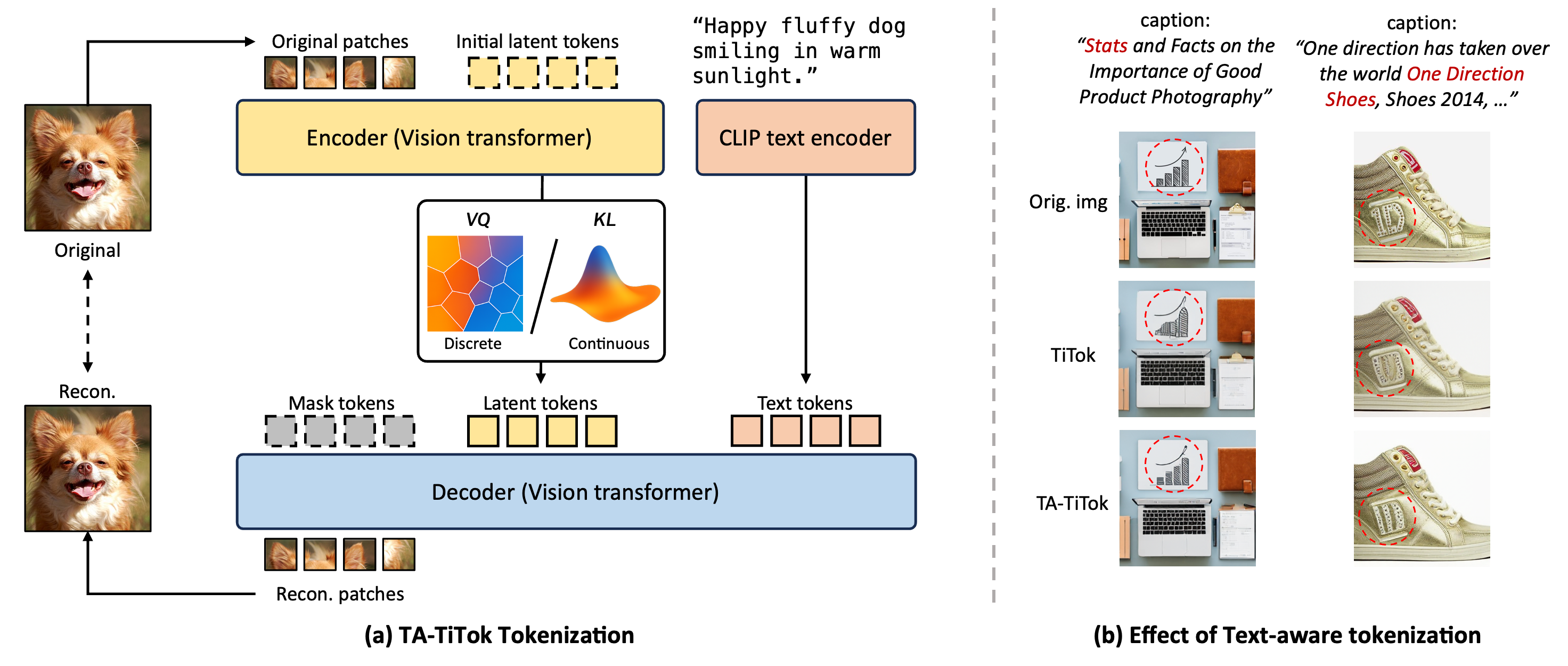}
    \vspace{-3mm}
    \caption{
    \textbf{Overview of \modelname (Text-Aware Transformer-based 1-Dimensional Tokenizer).}
    (a) \modelname introduces three key enhancements to TiTok~\cite{yu2024image}: First, an efficient one-stage training procedure replaces the need for a complex two-stage pipeline. Second, \modelname supports 1D tokens in both discrete (VQ) and continuous (KL) formats. Third, it incorporates textual information (using CLIP’s text encoder) during de-tokenization to improve semantic alignment with text captions.
    (b) A comparison of reconstruction results shows that \modelname achieves superior reconstruction quality over TiTok.
    }
    \vspace{-6mm}
    \label{fig:ta-titok}
\end{figure*}

\noindent \textbf{\basemodelname~\cite{yu2024image}} is a transformer-based, 1-dimensional Vector-Quantized (VQ) tokenizer that diverges from traditional 2D grid-based latent space tokenization, instead opting for a compact 1D representation that bypasses 2D spatial structure preservation.
Given an input image \(\mathbf{I} \in \mathbb{R}^{H \times W \times 3}\), the tokenization phase of \basemodelname involves downscaling the image by a factor of $f$, resulting in patches $\mathbf{P} \in \mathbb{R}^{\frac{H}{f} \times \frac{W}{f} \times D}$.
These patches are concatenated with a set of latent tokens $\mathbf{L} \in \mathbb{R}^{K \times D}$. The combined sequence is then passed through a Vision Transformer (ViT)~\cite{dosovitskiy2020image} encoder, $\mathtt{Enc}$, to generate embeddings. Only the embeddings corresponding to the latent tokens, $\mathbf{Z}_\text{1D} \in \mathbb{R}^{K \times D}$, are retained, forming a compact 1D latent representation. This representation is then quantized through a quantizer $\mathtt{Quant}$ by mapping it to the nearest codes in a learnable codebook.

In the de-tokenization phase, \basemodelname uses a sequence of mask tokens $\mathbf{M} \in \mathbb{R}^{\frac{H}{f} \times \frac{W}{f} \times D}$, which are concatenated with the quantized codes. The resulting sequence is processed by a ViT decoder, $\mathtt{Dec}$, to reconstruct the image \(\mathbf{\hat{I}}\). Formally, the tokenization and de-tokenization processes in \basemodelname can be represented as:
\begin{align}
    &[\_~; \mathbf{Z}_\text{1D}] = \mathtt{Enc}([\mathbf{P}; \mathbf{L}]), \\
    &[\_~; \mathbf{\hat{I}}] = \mathtt{Dec}([\mathtt{Quant}(\mathbf{Z}_\text{1D}); \mathbf{M}]),
\end{align}
where $[;]$ denotes concatenation along the sequence dimension, and $\_$ represents ignored tokens that are not used in subsequent operations.
 
\noindent \textbf{Masked Generative Models with Discrete Tokens}~\cite{chang2022maskgit,yu2023magvit} adapt the masked language modeling framework~\cite{devlin2018bert} for image generation. During training, a portion of image tokens is masked, and a bidirectional transformer predicts these tokens using the surrounding context. The model employs a classification head to select tokens from a predefined codebook~\cite{van2017neural} and uses cross-entropy loss for training.

During sampling, the model iteratively predicts tokens for masked positions, retaining high-confidence tokens while re-masking uncertain ones until all positions are filled~\cite{chang2022maskgit}. The completed sequence of tokens is then de-tokenized into pixel space to form the final image.

\noindent \textbf{Masked Generative Models with Continuous Tokens} maintain a conceptual similarity to discrete-token models but operate on continuous tokens, which reduces information loss from quantization. Recently, the diffusion loss~\cite{li2024autoregressive,fan2024fluid} was introduced, enabling these models to approximate the distribution of each image token independently. In this framework, Transformers generate a conditioning vector 
for each masked token, which is then input to a small multi-layer perceptron (MLP) that learns a denoising function~\cite{ho2020denoising} conditioned on 
it. This per-token conditioning and denoising allow the sampling process to directly apply to the probability distribution of each token~\cite{li2024autoregressive}.
Unlike discrete-token models, predicted tokens are masked randomly in each iteration.

%% file: sec/4_method.tex
\section{Method}
\label{sec:method}
In this section, we first present \modelname, a text-aware transformer-based 1-dimensional tokenizer (\secref{sec:ta-titok}), followed by \genmodelname, a family of text-to-image (T2I) masked generative models built upon \modelname (\secref{sec:maskgen}).

\subsection{\modelname: Text-aware 1D Tokenizer}
\label{sec:ta-titok}
\modelname is a novel text-aware image tokenizer that introduces three key enhancements to \basemodelname~\cite{yu2024image}. First, we develop an improved one-stage training approach; second, we extend \basemodelname to support both discrete and continuous tokens; and third, we incorporate textual information during de-tokenization to enhance semantic alignment. Each improvement is detailed below.

\noindent \textbf{Improved One-Stage Training Recipe.}
The recently proposed class-conditional masked generative model, MaskBit~\cite{weber2024maskbit}, introduced several techniques to enhance VQGAN~\cite{esser2021taming} training, two of which we incorporate in the training of our proposed \modelname.
First, MaskBit demonstrated that using ResNet50~\cite{he2016deep} for perceptual loss~\cite{johnson2016perceptual} yields richer features than the VGG network~\cite{simonyan2015very} used in LPIPS~\cite{zhang2018unreasonable}, thereby improving tokenizer training. Second, we strengthen the PatchGAN~\cite{esser2021taming} discriminator by replacing traditional average pooling with blur kernels~\cite{zhang2019making} and adding LeCAM regularization~\cite{tseng2021regularizing} during training.
Our experiments, consistent with MaskBit’s findings, confirm that these enhancements lead to improved image reconstruction quality through 1D tokens.

\noindent\textbf{Extending \basemodelname to Support Continuous Tokens.}
To maximize the efficiency of the \basemodelname framework in diffusion-based models, we extend its discrete 1D tokens to continuous 1D VAE representations.
Rather than using a quantizer $\mathtt{Quant}$ to map the 1D latents $\mathbf{Z}_\text{1D}$ to the nearest codebook entries, we model $\mathbf{Z}_\text{1D}$ as a Gaussian distribution and apply KL divergence regularization, resulting in a compact 1D VAE representation. This continuous representation retains the efficiency and structure of the \basemodelname framework, consistently improving reconstruction quality by avoiding the information loss associated with quantization.
Moreover, this KL variant of \basemodelname integrates seamlessly with diffusion models, serving as a drop-in replacement for standard 2D VAEs~\cite{rombach2022high,sdvae,li2024autoregressive}. 
In the supplementary materials (Tab. 13 in the Appendix), 
we validate this approach using the state-of-the-art image generation model MAR~\cite{li2024autoregressive} on ImageNet~\cite{deng2009imagenet}, where our modification achieves a significant reduction in training costs and an increase in inference speed, all while maintaining comparable performance.
This enhancement thus contributes to both the efficiency and flexibility of diffusion-based generation.
For clarity, we refer to both variants collectively as \basemodelname, using the labels VQ and KL to denote the discrete and continuous versions, respectively, in the following text.

\noindent \textbf{Text-aware De-tokenization.}
While \basemodelname effectively utilizes compact 1D tokens to capture richer semantic information than traditional 2D tokenizers, its primary focus remains on reconstructing low-level image details in the de-tokenization stage. Additionally, previous image tokenizers have largely overlooked the potential of textual information to enhance high-level semantic alignment, even when such information is available. This gap motivates us to introduce \modelname, a text-aware, transformer-based 1D tokenizer designed to improve alignment with textual descriptions.

As shown in~\figref{fig:ta-titok}(a), the tokenization stage in \modelname mirrors that of \basemodelname, transforming images into compact 1D tokens, either discrete or continuous. In the de-tokenization stage, however, \modelname incorporates text guidance by using text embeddings generated by a pre-trained vision-language model (\eg, CLIP's text encoder~\cite{radford2021learning}). These text embeddings are projected through a linear layer to align with the channel dimensions of our \modelname's ViT decoder, resulting in \(\mathbf{T} \in \mathbb{R}^{N \times D}\), where $N$ is the number of context tokens predefined by the vision-language model (\eg, 77 for CLIP's text encoder). This text embedding is then concatenated with the latent tokens $\mathbf{Z}_\text{1D}$ and the mask tokens $\mathbf{M}$ before passing through the decoder $\mathtt{Dec}$, yielding the reconstructed image \(\mathbf{\hat{I}}\).

Formally, the de-tokenization phase of the VQ variant of \modelname can be expressed as:
\begin{align}
    [\_~; \_~; \mathbf{\hat{I}}] = \mathtt{Dec}([\mathtt{Quant}(\mathbf{Z}_\text{1D}) ; \mathbf{T} ; \mathbf{M}]).
\end{align}
For the KL variant, the de-tokenization phase follows a similar formulation but omits the quantizer $\mathtt{Quant}$, as it operates on continuous representations directly.

\noindent \textbf{\modelname Design Discussion.}
Notably, \modelname incurs minimal additional computational cost compared to \basemodelname. By extending the de-tokenization sequence length by $N$ (\ie, the total number of tokens becomes $N+K$, where $N=77$ for CLIP’s text tokens and $K$ represents 32, 64, or 128 latent tokens), \modelname still requires fewer computations than typical 2D tokenizers~\cite{li2024autoregressive}, which utilize 256 tokens. This design enables \modelname to retain high efficiency while achieving reconstructions that closely align with text descriptions, effectively mitigating the information loss associated with compact 1D tokenization, as shown in~\figref{fig:ta-titok}(b).

The design of \modelname incorporates text tokens exclusively into the tokenizer decoder with minimal modifications. To validate this approach, we also experimented with adding text tokens to both the encoder and decoder. The results showed similar performance to the decoder-only approach, suggesting that incorporating textual information during de-tokenization is sufficient for capturing high-level semantic information. Based on these findings, we adopt the simpler decoder-only design. 
For additional details, see Tab.10 in the Appendix.

\begin{figure*}[!ht]
    \centering
    \includegraphics[width=0.96\linewidth]{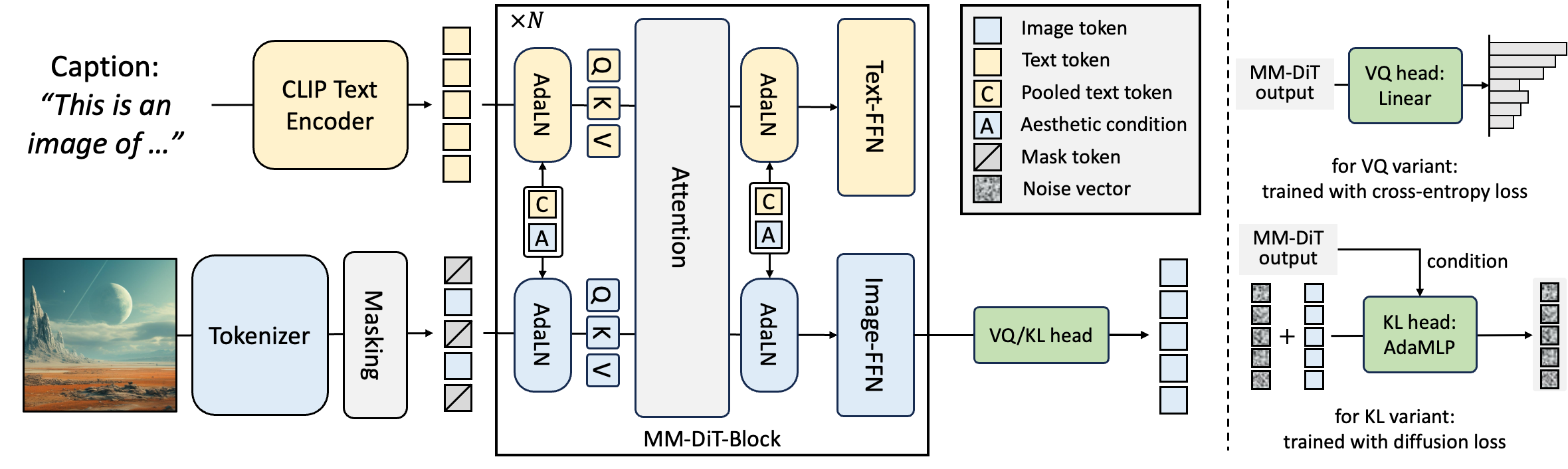}
    \vspace{-2mm}
    \caption{
    \textbf{Overview of \genmodelname.}
    \genmodelname is a family of text-to-image masked generative models that supports both discrete (VQ variant) and continuous (KL variant) token representations. For discrete tokens, \genmodelname is trained with cross-entropy loss~\cite{chang2022maskgit}, while for continuous tokens, it employs diffusion loss~\cite{li2024autoregressive}.
    The architecture is designed by concatenating text conditions with \modelname's latent tokens (both masked and unmasked) and feeding them into Diffusion Transformer blocks~\cite{peebles2023scalable}, with separate adaptive LayerNorms (adaLN), linear projections, and feedforward networks (FFN) for text and image modalities, following MM-DiT~\cite{esser2024scaling}. Additionally, aesthetic scores are incorporated as conditioning signals via adaLN. To encode captions, MaskGen uses the CLIP text encoder~\cite{radford2021learning} instead of the more resource-intensive T5-XXL~\cite{raffel2020exploring}, making it more accessible to research groups with limited computational resources. 
    }
    \label{fig:maskgen}
\end{figure*}

\subsection{\genmodelname: T2I Masked Generative Model}
\label{sec:maskgen}

To fully leverage \modelname's capabilities, we propose \genmodelname, a family of text-to-image (T2I) masked generative models. As illustrated in~\figref{fig:maskgen}, \genmodelname utilizes \modelname to tokenize images into tokens and a CLIP~\cite{radford2021learning} text encoder to extract both global and pooled text embeddings. Inspired by~\cite{esser2024scaling}, we concatenate the global text embedding with the image tokens, feeding this combined sequence into multi-modal Diffusion Transformer (MM-DiT) blocks~\cite{esser2024scaling,peebles2023scalable} for attention operations. To accommodate the distinct properties of text and image embeddings, we apply separate adaptive LayerNorm (adaLN~\cite{ba2016layer,peebles2023scalable}) layers, where the scale and shift parameters are computed based on the pooled text embedding.
We note that while a more powerful text encoder, such as T5-XXL~\cite{raffel2020exploring}, could be directly integrated into the pipeline to enhance performance, as demonstrated in studies like~\cite{saharia2022photorealistic,chang2023muse,esser2024scaling}, we choose CLIP for its computational efficiency and reduced storage demands, making \genmodelname more accessible in resource-constrained settings.

We also incorporate aesthetic scores~\cite{schuhmann2022laion} as another condition by projecting them into sinusoidal embeddings and appending them to the pooled text embedding. This feature provides an extra layer of control over the image quality and style during sampling, making the generation process more adaptable and flexible.

\genmodelname is a versatile framework that accommodates both discrete and continuous tokens produced by \modelname. For discrete tokens, \genmodelname is trained with a cross-entropy loss, as in~\cite{chang2022maskgit,chang2023muse}, to predict the correct one-hot codebook index for masked tokens. In the case of continuous tokens, \genmodelname leverages the recently proposed diffusion loss~\cite{li2024autoregressive}, applying a small MLP to directly approximate the distribution of each masked token. This adaptability makes \genmodelname capable of handling various token types with ease. Additionally, thanks to the compact 1D token sequence produced by \modelname, \genmodelname is highly efficient, reducing training costs and enhancing sampling speed by minimizing token count. Together, these features help \genmodelname to democratize access to efficient, high-performance masked generative models for text-to-image generation.

%% file: sec/5_experiment.tex
\section{Experimental Results}
\label{sec:experiment}

\subsection{Implementation Details}
\label{sec:details}

\noindent \textbf{\modelname Model Variants.} We present three variants of our \modelname, each containing $K$ = 32, 64, or 128 1D latent tokens, following the architecture of \basemodelname. Both the tokenizer and de-tokenizer utilize a patch size of $f = 16$. For the VQ variant, the codebook is configured with 8192 entries, where each entry is a vector with 64 channels. For the KL variant, we use a continuous embedding with 16 channels, following MAR~\cite{li2024autoregressive}.
For the encoder ($\mathtt{Enc}$) and decoder ($\mathtt{Dec}$), we find that increasing the size of $\mathtt{Dec}$ to ViT-L~\cite{dosovitskiy2020image} is beneficial when training on large-scale datasets across all variants. For $\mathtt{Enc}$, we adopt ViT-B for all variants except the KL variant with 128 tokens, where ViT-S is sufficient.

\input{sec/tables/maskgen_configurations}

\noindent \textbf{\genmodelname Model Variants.} We introduce two variants of \genmodelname: \genmodelname-L (568M parameters) and \genmodelname-XL (1.1B parameters), with configurations detailed in~\tabref{tab:maskgen_configuration}. For continuous token processing in \genmodelname, we incorporate an additional DiffLoss MLP~\cite{li2024autoregressive}, comprising 8 MLP layers with channel sizes aligned to the transformer's, adding an extra 44M and 69M parameters for \genmodelname-L and \genmodelname-XL, respectively. To offset the additional training and inference cost introduced by the DiffLoss MLP, we use 128 tokens for the discrete \genmodelname variant and 32 tokens for the continuous \genmodelname variant. 

\noindent \textbf{Training Data.}
For training \modelname and \genmodelname, we utilized various open-source datasets: DataComp-1B~\cite{gadre2024datacomp}, CC12M~\cite{changpinyo2021conceptual}, LAION-aesthetic~\cite{laion-aesthetic}, LAION-art~\cite{laion-art}, LAION-pop~\cite{laion-pop}, JourneyDB~\cite{sun2024journeydb}, and DALLE3-1M~\cite{Egan_Dalle3_1_Million_2024}. 
\modelname is trained exclusively with DataComp-1B. \genmodelname undergoes a two-stage training process: pre-training for image-text alignment and fine-tuning using aesthetic images.
Details regarding the dataset preparation and recaptioning process are provided in the Appendix.

\noindent \textbf{Evaluation Metrics.}
Our evaluation pipeline closely follows prior works~\cite{yu2022scaling,chang2023muse}.
The images are generated without rejection sampling, and classifier-free guidance~\cite{ho2022classifier} is used to enhance generation quality. 
Unless specified otherwise, \genmodelname uses 16 and 32 sampling steps for VQ and KL variants, respectively.
For \modelname, we measure reconstruction quality using reconstruction FID (rFID)~\cite{heusel2017gans} and inception score (IS)~\cite{salimans2016improved} on ImageNet~\cite{deng2009imagenet} validation set. 
For \genmodelname's text-to-image generation capabilities, we utilize FID on MJHQ~\cite{li2024playground} to assess aesthetic quality, and GenEval~\cite{ghosh2024geneval} score to measure the alignment between text prompts and their corresponding generated images.

\input{sec/tables/ablation_one-stage}

\subsection{Optimized Image Tokenization with \modelname}
\label{sec:ablation}

\noindent\textbf{Improved One-Stage Training Recipe.}
\tabref{tab:ablation_one-stage} summarizes the performance gains of our improved one-stage training recipe over the original schemes in~\cite{yu2024image}. As observed, the adopted one-stage training significantly outperforms the original, achieving an rFID improvement of 2.72.

\begin{figure}
    \centering
    \includegraphics[width=0.96\columnwidth]{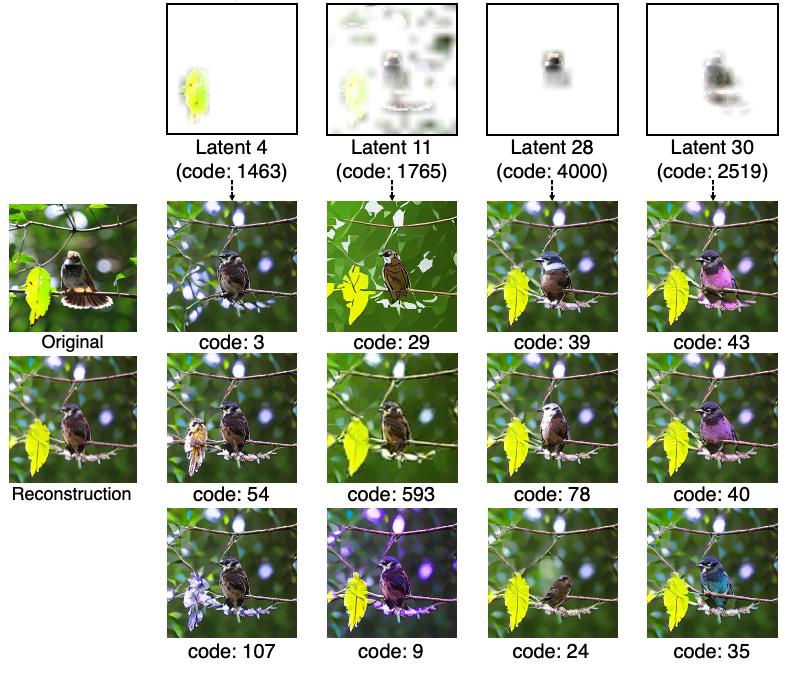}
    \vspace{-3mm}
    \caption{
    \textbf{Visualization of Latent Token Attention Map and Latent Code Swapping.} The results are from VQ variant of \modelname with 32 tokens. Each latent token attends to prominent semantic and swapping the code leads to appearance changes in the corresponding semantic entity that the latent token focuses on.
    }
    \vspace{-5mm}
    \label{fig:token_swap}
\end{figure}

\noindent \textbf{Latent Token Attention Maps Reveal Semantic Focus.} 
In the first row of \figref{fig:token_swap}, we overlay the attention maps of each latent token onto the image, where attention is computed from the encoder's final layer. 
We observe that each latent token focuses on semantically meaningful regions within the image, rather than specific spatial locations or grid cells as in 2D tokenization. 
For instance, individual latents capture distinct semantic elements such as leaf, the bird's head, and the bird's torso. 
Notably, we find that at least one latent token (\eg, Latent 11) attends to the entire image, and subsequent experiments reveal that it captures the overall visual style.

\noindent \textbf{Semantic Manipulation Through Latent Code Swapping.}
\figref{fig:token_swap} also presents reconstruction results when we swap the code for latent tokens. 
Interestingly, swapping the code leads to changes in the semantic element corresponding to each latent token, with appearance changes ranging from low-level visual features (\eg, color changes) to high-level semantics (\eg, new objects, object removal, pose changes). 
For instance, swapping the code for the 4th latent token, which focuses on the leaf, leads to the removal of the leaf or its transformation into a bird or flower.
Swapping for the 28th latent token results to the change in bird's head in terms of its spices or pose, and for 30th latent token swapping results to color change of bird's torso.
Unlike other latent tokens, swapping the 11th latent token focusing on the overall image leads to the change of overall style of the image. 
For instance, swapping to code 29 leads to the whole scene being redrawn in a flat, vector-graphic style; swapping to code 593 softens high-frequency details; swapping to code 9 produces a purple-tinted stylistic transformation.

\input{sec/tables/abltation_text-aware}

\noindent\textbf{Effect of Text-Aware De-Tokenization.}
We evaluate the impact of text-aware de-tokenization (\modelname) in~\tabref{tab:ablation_text-aware}. For consistency, all models are trained using our improved one-stage training recipe. Tokenizers are trained on the DataComp dataset~\cite{gadre2024datacomp} and evaluated in a zero-shot setting on the ImageNet validation set~\cite{deng2009imagenet}, where the caption is simply represented as ``A photo of \emph{class}'' without any prompt engineering.
We compare two architectures—VQ and KL variants—and vary the token count between 32, 64, and 128. As shown in the table, continuous tokens consistently outperform discrete tokens, aligning with findings in~\cite{esser2021taming}. Additionally, \modelname consistently outperforms TiTok (the non-text-aware variant) across all configurations. Notably, the performance gain is most pronounced with a smaller number of tokens (\eg, 32) and with discrete tokens. 
We attribute this to the text embeddings supplementing high-level semantic information overlooked by latent tokens. 
Consequently, the improvement is more substantial with fewer tokens and vector-quantized tokens, where latent tokens experience greater information loss.

\input{sec/tables/t2i_mjhq}

\input{sec/tables/t2i_geneval}

\subsection{Text-to-Image Generation with \genmodelname}
\label{sec:main_results}

\noindent\textbf{MJHQ-30K.}
We report zero-shot text-to-image generation results on MJHQ-30K~\cite{li2024playground} in~\tabref{tab:t2i_mjhq}. \genmodelname-XL (discrete tokens) achieves a significantly better FID than recent autoregressive models like LlamaGen~\cite{sun2024autoregressive} (7.51 \vs 25.59) and Show-o~\cite{xie2024show} (7.51 \vs 14.99), both using VQ tokenizers. It also offers 18.5$\times$ higher inference throughput than Show-o, with \genmodelname-L being 30.3$\times$ faster, though with a slight performance drop. \genmodelname demonstrates impressive resource efficiency, with \genmodelname-L completing training in 20.0 8-A100 days and \genmodelname-XL in 35.0 8-A100 days.

For continuous tokens, \genmodelname delivers competitive results against recent diffusion models. With just 32 tokens, \genmodelname-L outperforms PixArt-$\alpha$\cite{chen2024pixartalpha} (630M) (7.24 \vs 9.85), offering 1.4$\times$ faster inference throughput while using fewer parameters and requiring less than one-fifth of the training resources (18.5 \vs 94.1 8-A100 days). \genmodelname-XL further improves the FID score to 6.53, surpassing SDXL\cite{podell2023sdxl} (6.53 \vs 8.76), a 2.6B-parameter model trained on high-quality private data, despite \genmodelname-XL being trained exclusively on open data for only 30.5 8-A100 days.

\noindent\textbf{GenEval.}
\tabref{tab:t2i_geneval} summaries the zero-shot text-to-image generation results on GenEval~\cite{ghosh2024geneval}.
Using discrete tokens, \genmodelname-L (568M) achieves an overall score of 0.53, significantly outperforming the recent autoregressive model LlamaGen~\cite{sun2024autoregressive} by 0.21 and performing on par with the larger Show-o~\cite{xie2024show} (1.3B).
Moreover, the larger \genmodelname-XL achieves the highest overall score on the benchmark, with a score of 0.57. This result notably surpasses SDXL~\cite{podell2023sdxl}, a 2.6B-parameter model (2.36$\times$ larger than \genmodelname-XL) trained on proprietary data.
Meanwhile, our \genmodelname-XL with continuous tokens also achieves an overall score of 0.55, comparable to recent diffusion models~\cite{rombach2022high}, but with much lower training costs and exclusively trained on open data.

\input{sec/tables/ablation_tech-designs}

\noindent\textbf{Effect of Text-Aware Design and Aesthetic Score Conditioning.}
For efficient ablation studies, we use the continuous version of \genmodelname-L. As shown in \tabref{tab:ablation_tech-designs}, adopting \modelname significantly improves generation quality (Row 1 \vs Row 3). Additionally, incorporating aesthetic score conditioning further enhances performance (Row 3 \vs Row 4).
Moreover, comparing Row 2 with other configurations suggests that while aesthetic score conditioning helps \genmodelname better control image fidelity, its impact is less significant than that of the text-aware tokenizer \modelname.

%% file: sec/tables/maskgen_configurations.tex
\begin{table}[]
\centering
\small
\tablestyle{2.0pt}{1.0}
\scalebox{1.2}{
\begin{tabular}{l|ccccc}
model & depth & width & mlp & heads & \#params \\ \shline
\genmodelname-L & 16 & 1024 & 4096 & 16 & 568M \\
\genmodelname-XL & 20 & 1280 & 5120 & 16 & 1.1B 
\end{tabular}
}
\caption{
\textbf{Architecture Configuration of \genmodelname.}
Following prior work, we scale up MM-DiT blocks across two configurations.
}
\vspace{-4ex}
\label{tab:maskgen_configuration}
\end{table}

%% file: sec/tables/ablation_one-stage.tex
\begin{table}
\centering
\small
\tablestyle{2.0pt}{1.0}
\scalebox{1.2}{
\begin{tabular}{lcccccc}
tokenizer & arch & training setting & \#tokens & rFID$\downarrow$ & IS$\uparrow$ \\
\shline
\multirow{1}{*}{\basemodelname~\cite{yu2024image}} & \multirow{1}{*}{VQ} & 1-stage~\cite{esser2021taming} & \multirow{1}{*}{64} & 5.15 & 120.5  \\
\hline
\basemodelname & VQ & our 1-stage & 64 & 2.43 & 179.3 \\
\end{tabular}
}
\vspace{-2mm}
\caption{\textbf{Ablation on Improved One-Stage Training Recipe.} Both models are trained and evaluated on ImageNet.
}
\vspace{-2mm}
\label{tab:ablation_one-stage}
\end{table}

%% file: sec/tables/abltation_text-aware.tex
\begin{table}
\centering
\small
\tablestyle{2.0pt}{1.0}
\scalebox{1.1}{
\begin{tabular}{ccc|cc|cc}
\multirow{2}{*}{arch} & \multicolumn{2}{c|}{tokens} & \multicolumn{2}{c|}{\basemodelname} & \multicolumn{2}{c}{\modelname} \\
 & \# & c & rFID$\downarrow$ & IS$\uparrow$ & rFID$\downarrow$ & IS$\uparrow$ \\ \shline
\multirow{3}{*}{VQ} & 32 & - & 7.72 & 98.3 & 3.95 \textcolor{blue}{(-3.77)} & 219.6 \textcolor{blue}{(+121.3)} \\
& 64 & - & 4.25 & 138.0 & 2.43 \textcolor{blue}{(-1.82)} & 218.8 \textcolor{blue}{(+80.8)} \\
& 128 & - & 2.63 & 168.1 & 1.53 \textcolor{blue}{(-1.10)} & 222.8 \textcolor{blue}{(+54.7)} \\
\hline
\multirow{3}{*}{KL} & 32 & 16 & 2.56 & 171.7 & 1.53 \textcolor{blue}{(-1.03)} & 222.0 \textcolor{blue}{(+50.3)} \\
& 64 & 16 & 1.64 & 198.0 & 1.47 \textcolor{blue}{(-0.17)} & 220.7 \textcolor{blue}{(+22.7)} \\
& 128 & 16 & 1.02 & 209.7 & 0.90 \textcolor{blue}{(-0.12)} & 227.7 \textcolor{blue}{(+18.0)}
\end{tabular}
}
\vspace{-3mm}
\caption{\textbf{Ablation on Text-Aware De-tokenization Design.}
All models are trained on DataComp with our improved one-stage recipe and evaluated in a zero-shot setting on the ImageNet validation set. Relative improvements for \modelname are highlighted in blue. \#: Number of tokens. c: Channels of continuous tokens.
}
\vspace{-3mm}
\label{tab:ablation_text-aware}
\vspace{-2ex}
\end{table}

%% file: sec/tables/t2i_mjhq.tex
\begin{table*}[!t]
\small
\centering
\tablestyle{2.0pt}{1.0}
\scalebox{1.2}{
\begin{tabular}{lc|clcc|ccc|c}
tokenizer & arch & type & generator & \#params & resolution & open-data  & T$\downarrow$ & I$\uparrow$ & FID$\downarrow$ \\ \shline
VQGAN~\cite{sun2024autoregressive} & VQ & AR & LlamaGen~\cite{sun2024autoregressive} & 775M & 512 $\times$ 512 & \xmark & - & - & 25.59 \\ 
MAGVIT-v2~\cite{xie2024show} & VQ & AR & Show-o~\cite{xie2024show} & 1.3B & 256 $\times$ 256 & \cmark & - & 1.0 & 14.99 \\
\hline
\modelname & VQ & Mask. & \genmodelname-L (ours) & 568M & 256 $\times$ 256 & \cmark & 20.0 & 30.3 & 7.74 \\
\modelname & VQ & Mask. & \genmodelname-XL (ours) & 1.1B & 256 $\times$ 256 & \cmark & 35.0 & 18.5 & 7.51 \\
\hline
\hline
VAE~\cite{rombach2022high} & KL & Diff. & Stable-Diffusion-2.1~\cite{rombach2022high} & 860M & 768 $\times$ 768 & \cmark & 1041.6 & - & 26.96 \\
VAE~\cite{rombach2022high} & KL & Diff. & PixArt-$\alpha$~\cite{chen2024pixartalpha} & 630M & 256 $\times$ 256 & \xmark & 94.1 & 7.9 & 9.85 \\
VAE~\cite{podell2023sdxl} & KL & Diff. & SDXL~\cite{podell2023sdxl} & 2.6B & 1024 $\times$ 1024 & \xmark & - & - & 8.76 \\
\hline
\modelname & KL & Mask. & \genmodelname-L (ours) & 568M + 44M & 256 $\times$ 256 & \cmark & 18.5 & 11.1 & 7.24 \\
\modelname & KL & Mask. & \genmodelname-XL (ours) & 1.1B + 69M & 256 $\times$ 256 & \cmark & 30.5 & 9.1 & 6.53 \\
\end{tabular}
}
\vspace{-3mm}
\caption{
\textbf{Zero-Shot Text-to-Image Generation Results on MJHQ-30K.}
Comparison of \genmodelname with state-of-the-art \textit{open-weight} models.
``VQ'' denotes discrete tokenizers and ``KL'' stands for continuous tokenizers.
``type'' indicates the generative model type, where ``AR'', ``Diff.'' and ``Mask.'' refer to autoregressive models, diffusion models and masked transformer models, respectively.
T: Generator training cost, measured in 8 A100 days using float16 precision. I: Generator inference throughput, measured in samples per second on a single A100 with batch size 64 using float16 precision. We compare inference throughput with methods using the same resolution.
}
\vspace{-2mm}
\label{tab:t2i_mjhq}
\end{table*}

%% file: sec/tables/t2i_geneval.tex
\begin{table*}[!t]
\small
\centering
\tablestyle{2.0pt}{1.0}
\scalebox{1.12}{
\begin{tabular}{lc|lc|c|cccccc|c}
tokenizer & arch & generator & \#params & open-data & S. Obj. & T. Obj. & Count. & Colors & Position & C. Attri. & Overall$\uparrow$ \\ \shline
VQGAN~\cite{sun2024autoregressive} & VQ & LlamaGen~\cite{sun2024autoregressive} & 775M & \xmark & 0.71 & 0.34 & 0.21 & 0.58 & 0.07 & 0.04 & 0.32 \\
MAGVIT-v2~\cite{xie2024show} & VQ & Show-o~\cite{xie2024show} & 1.3B & \cmark & 0.95 & 0.52 & 0.49 & 0.82 & 0.11 & 0.28 & 0.53 \\
\hline
\modelname & VQ & \genmodelname-L (ours) & 568M & \cmark & 0.98 & 0.57 & 0.46 & 0.80 & 0.11 & 0.25 & 0.53 \\
\modelname & VQ & \genmodelname-XL (ours) & 1.1B & \cmark & 0.99 & 0.61 & 0.55 & 0.81 & 0.13 & 0.31 & 0.57 \\
\hline
\hline
VAE~\cite{rombach2022high} & KL & Stable-Diffusion-1.5~\cite{rombach2022high} & 860M & \cmark & 0.97 & 0.38 & 0.35 & 0.76 & 0.04 & 0.06 & 0.43 \\
VAE~\cite{rombach2022high} & KL & PixArt-$\alpha$~\cite{chen2024pixartalpha} & 630M & \xmark & 0.96 &  0.49&  0.47&  0.79&  0.06&  0.11& 0.48 \\
VAE~\cite{rombach2022high} & KL & Stable-Diffusion-2.1~\cite{rombach2022high} & 860M & \cmark & 0.98 & 0.51 & 0.44 & 0.85 & 0.07 & 0.17 & 0.50 \\
VAE~\cite{podell2023sdxl} & KL & SDXL~\cite{podell2023sdxl} & 2.6B & \xmark & 0.98 & 0.74 & 0.39 & 0.85 & 0.15 & 0.23 & 0.55 \\
\hline
\modelname & KL & \genmodelname-L (ours) & 568M + 44M & \cmark & 0.99 & 0.57 & 0.36 & 0.80 & 0.11 & 0.29 & 0.52 \\
\modelname & KL & \genmodelname-XL (ours) & 1.1B + 69M & \cmark & 0.99 & 0.58 & 0.47 & 0.77 & 0.13 & 0.34 & 0.55
\end{tabular}
}
\vspace{-3mm}
\caption{
\textbf{Zero-Shot Text-to-Image Generation Results on GenEval.}
Comparison of \genmodelname with state-of-the-art \textit{open-weight} models.
}
\vspace{-6mm}
\label{tab:t2i_geneval}
\end{table*}

%% file: sec/tables/ablation_tech-designs.tex
\begin{table}
\centering
\small
\tablestyle{2.0pt}{1.0}
\scalebox{1.1}{
\begin{tabular}{lc|c|c|c|c}
     &           &          &          & MJHQ-30K & GenEval \\
tokenizer & arch & generator & aesthetic & FID$\downarrow$ & Overall$\uparrow$ \\ \shline
\basemodelname & \multirow{4}{*}{KL} & \multirow{4}{*}{\genmodelname-L} & \xmark & 9.82 & 0.47 \\
\basemodelname &  &  & \cmark & 8.50 & 0.49 \\
\modelname &  &  & \xmark & 8.13 & 0.51 \\
\modelname &  &  & \cmark & 7.24 & 0.52 \\
\end{tabular}
}
\vspace{-3mm}
\caption{
\textbf{Ablation of Text-Aware Design and Aesthetic Score Condition.}
Incorporating aesthetic scores and exploiting \modelname enhance generation quality.
}
\vspace{-6mm}
\label{tab:ablation_tech-designs}
\end{table}

%% file: sec/6_conclusion.tex
\section{Conclusion}
We introduced \modelname, a text-aware 1D tokenizer that enhances semantic alignment, and \genmodelname, a versatile masked generative model supporting discrete and continuous tokens. With compact 1D tokenization, \genmodelname lowers training costs and accelerates sampling, enabling efficient, high-quality generation. Leveraging public data, it broadens access to high-performance generative models. All code and model weights is released to promote future research.

%% file: sec/X_suppl.tex
\clearpage
\setcounter{page}{1}
\maketitlesupplementary

\renewcommand{\thesection}{\Alph{section}}
\setcounter{section}{0}

\section*{Appendix}
\label{sec:appendix}

We provide additional information in the supplementary material, as outlined below:

\begin{itemize}
\item Sec.~\ref{sec:dataset_preparation}: Further implementation details, including dataset filtering, recaptioning, and MaskGen training hyperparameters.
\item Sec.~\ref{sec:ablation_tatitok}: Ablation studies on the type and place of text guidance in \modelname.
\item Sec.~\ref{sec:ablation_maskgen}: Ablation studies on the number of tokens and aesthetic score condition in \genmodelname.
\item Sec.~\ref{sec:t2i_comp}: Comparisons between \genmodelname using discrete and continuous tokens.
\item Sec.~\ref{sec:coco}: Additional zero-shot text-to-image generation results on COCO validation set.
\item Sec.~\ref{sec:mar_imagenet}: Results demonstrating the performance of our KL variant of TiTok on ImageNet.
\item Sec.~\ref{sec:more_swap_vis}: More qualitative examples for latent code swapping with \modelname.
\item Sec.~\ref{sec:more_vis}: More qualitative examples generated by \genmodelname.
\item Sec.~\ref{sec:limitations}: Discussions on limitations and future work of \modelname and \genmodelname.
\end{itemize}

\section{More Implementation Details}
\label{sec:dataset_preparation}
\input{sec/tables/dataset_detail}

\noindent \textbf{Dataset Filtering.} To prepare training data, we applied three filtering criteria: resolution, aesthetic, and watermark filtering. Details of applied filtering criteria and the total size of the dataset after filtering are presented in~\tabref{tab:dataset_detail}.
Resolution filtering was applied to all datasets during the training of both \modelname and \genmodelname. This filtering ensured that the longer side of each image exceeded 256 pixels and maintained an aspect ratio below 2. 
For \genmodelname training, we implemented aesthetic filtering using the LAION-aesthetic-v2 predictor~\cite{laion-aesthetic-predictor} to retain only high-quality images. Images with scores above 5.0 were kept during the pre-training stage, while a stricter threshold of 6.0 was applied during fine-tuning to ensure even higher quality. 
Additionally, we employed watermark filtering for \genmodelname using the LAION-WatermarkDetector~\cite{laion-watermark-detector}, removing images with watermark probability exceeding 0.5 to prevent unwanted watermark artifacts in generated images. Synthetic datasets such as JourneyDB~\cite{sun2024journeydb} and DALLE3-1M~\cite{Egan_Dalle3_1_Million_2024} were exempted from these filtering processes as they inherently met our high resolution and quality standards.

\begin{figure*}[t!]
    \centering
    \includegraphics[width=0.96\linewidth]{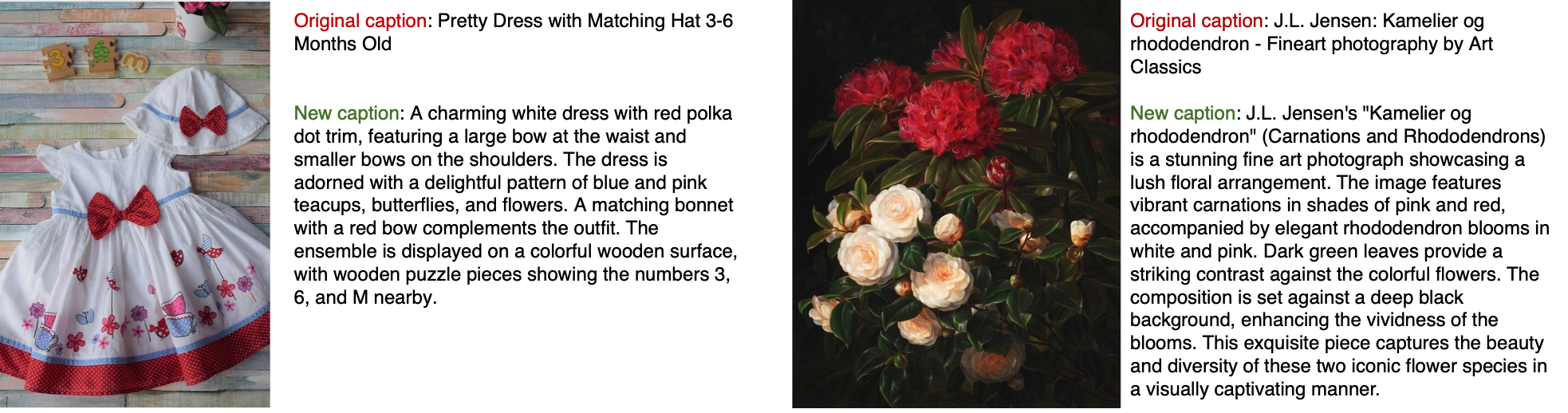}
    \caption{
    \textbf{Re-captioning Results.}
    Captions augmented by Molmo~\cite{deitke2024molmo} offer richer details and improved alignment with image content.
    }
    \vspace{-4mm}
    \label{fig:recap}
\end{figure*}

\noindent\textbf{Dataset Recaptioning.} To improve the text quality of DataComp~\cite{gadre2024datacomp}, LAION-art~\cite{laion-art}, and LAION-pop~\cite{laion-pop}, we utilize state-of-the-art vision-language models, Molmo-7B-D-0924~\cite{deitke2024molmo}, to enhance captions based on both the image and its original caption. Specifically, we randomly sample one of four prompts as shown in~\figref{fig:prompts} to generate updated captions. Since the augmented captions are often significantly longer than the original ones and frequently start with similar patterns (\eg, ``The image depicts/displays/showcases/shows/features...''), we apply prefix filtering to remove these repetitive prefixes, preventing information leakage. During training, we further address this by employing a 95:5 ratio, randomly sampling between augmented and original captions to ensure balanced learning, following~\cite{betker2023improving}. A few recaption results are shown in~\figref{fig:recap}, highlighting how the augmented captions provide richer details and align better with the image content.

\noindent \textbf{Training.} 
We adhere strictly to the hyperparameters used to train \basemodelname across all \modelname variants. Specifically, \modelname is trained with a batch size of 1024 for 1 epoch (650k steps) on the filtered DataComp dataset, using a maximum learning rate of 1e-4 and a cosine learning rate schedule. For \genmodelname with discrete tokens, we employ a batch size of 4096, leveraging weight tying~\cite{press2016using} to stabilize training, with a cosine learning rate schedule and a maximum learning rate of 4 × 10\textsuperscript{-4}. For \genmodelname with continuous tokens, to accommodate the diffusion loss, we use a constant learning rate schedule with a maximum rate of 1 × 10\textsuperscript{-4} and a batch size of 2048. Masked tokens are sampled by randomly selecting the masking rate from [0, 1] on a cosine schedule, following MaskGIT~\cite{chang2022maskgit}, and text conditioning is randomly dropped with a 0.1 probability to enable classifier-free guidance~\cite{ho2022classifier}. 
\tabref{tab:training_hparams} provides the complete list of hyper-parameters used for training \genmodelname with both discrete and continuous tokenizers.

\input{sec/tables/prompts}

\input{sec/tables/training_hparams}

\input{sec/tables/ablation_text-type}
\input{sec/tables/ablation_text-place}
\section{Ablation Studies for \modelname}
\label{sec:ablation_tatitok}
\noindent\textbf{Text Guidance Type in \modelname.}
In our text-aware de-tokenization design, we can use either the numerical IDs from the CLIP text tokenizer or the embeddings from the CLIP text encoder. We ablate this design choice in~\tabref{tab:ablation_text-type}, where the latter option yields a marginal improvement.

\noindent \textbf{Text Guidance Place in \modelname}
In the design of \modelname, we only incorporate the text guidance (\ie, the text tokens from CLIP) into the tokenizer decoder to better capture high-level semantics and align with textual descriptions during both reconstruction and generation. 
In this study, we investigate whether injecting text guidance into both the encoder and decoder of \modelname can further enhance the quality of the encoded latent tokens. This evaluation is conducted on the ImageNet validation set~\cite{deng2009imagenet} using reconstruction metrics in a zero-shot setting, where the caption is represented as ``A photo of \emph{class}'' without any prompt engineering.
As shown in~\tabref{tab:ablation_text-place}, injecting textual guidance into both the encoder and decoder has negligible impact on reconstruction quality. This finding suggests that incorporating text guidance in the decoder alone is sufficient to provide semantic information to the model.

\section{Ablation Studies for \genmodelname}
\label{sec:ablation_maskgen}
\noindent\textbf{Experimental Setup.}
For efficient ablation studies, we use the discrete version of \genmodelname to analyze the impact of token count. Performance is evaluated using the FID metric on MJHQ-30K~\cite{li2024playground} and the overall GenEval~\cite{ghosh2024geneval} score. Additionally, we provide visual comparisons to illustrate the effect of aesthetic score conditioning during sampling.

\noindent\textbf{Number of Tokens.}
Tab.~\ref{tab:ablation_token_count} presents an ablation study on the number of tokens used for text-to-image generation with \genmodelname. As observed, increasing the token count improves generation quality but comes at the expense of longer training times and slower inference speeds.

\input{sec/tables/ablation_token_count}

\begin{figure}
    \centering
    \includegraphics[width=0.96\columnwidth]{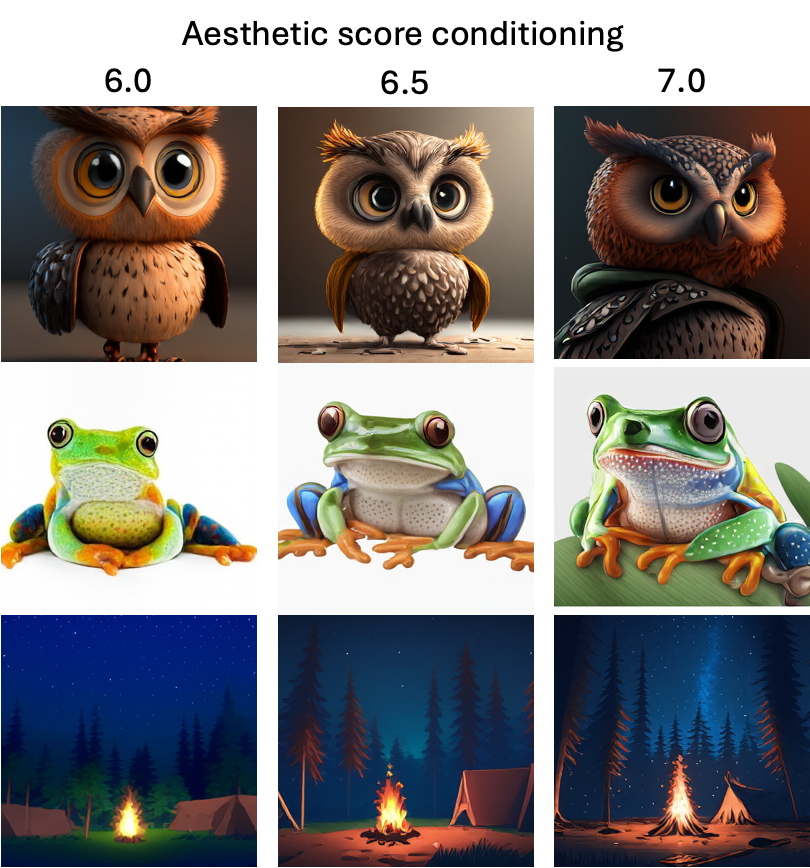}
    \vspace{-2mm}
    \caption{
    \textbf{Generated Images with Varying Aesthetic Score Conditioning.} Conditioning on higher aesthetic scores produces generated images with enhanced fine-grained details.
    }
    \vspace{-2mm}
    \label{fig:aes}
\end{figure}

\noindent\textbf{Aesthetic Score Conditioning.}
~\figref{fig:aes} visualizes images generated with different aesthetic scores while keeping other hyperparameters and prompts constant. The results indicate a strong correlation between higher aesthetic scores and enhanced dramatic lighting and fine-grained details. For instance, in the third row, a higher aesthetic score yields richer depictions of trees and stars in the night sky, whereas a lower score results in simpler representations. This enables precise control over image generation based on user preferences.

\section{Comparisons Between \genmodelname Using Discrete and Continuous Tokens}
\label{sec:t2i_comp}
\begin{figure*}[!ht]
    \centering
    \includegraphics[width=\linewidth]{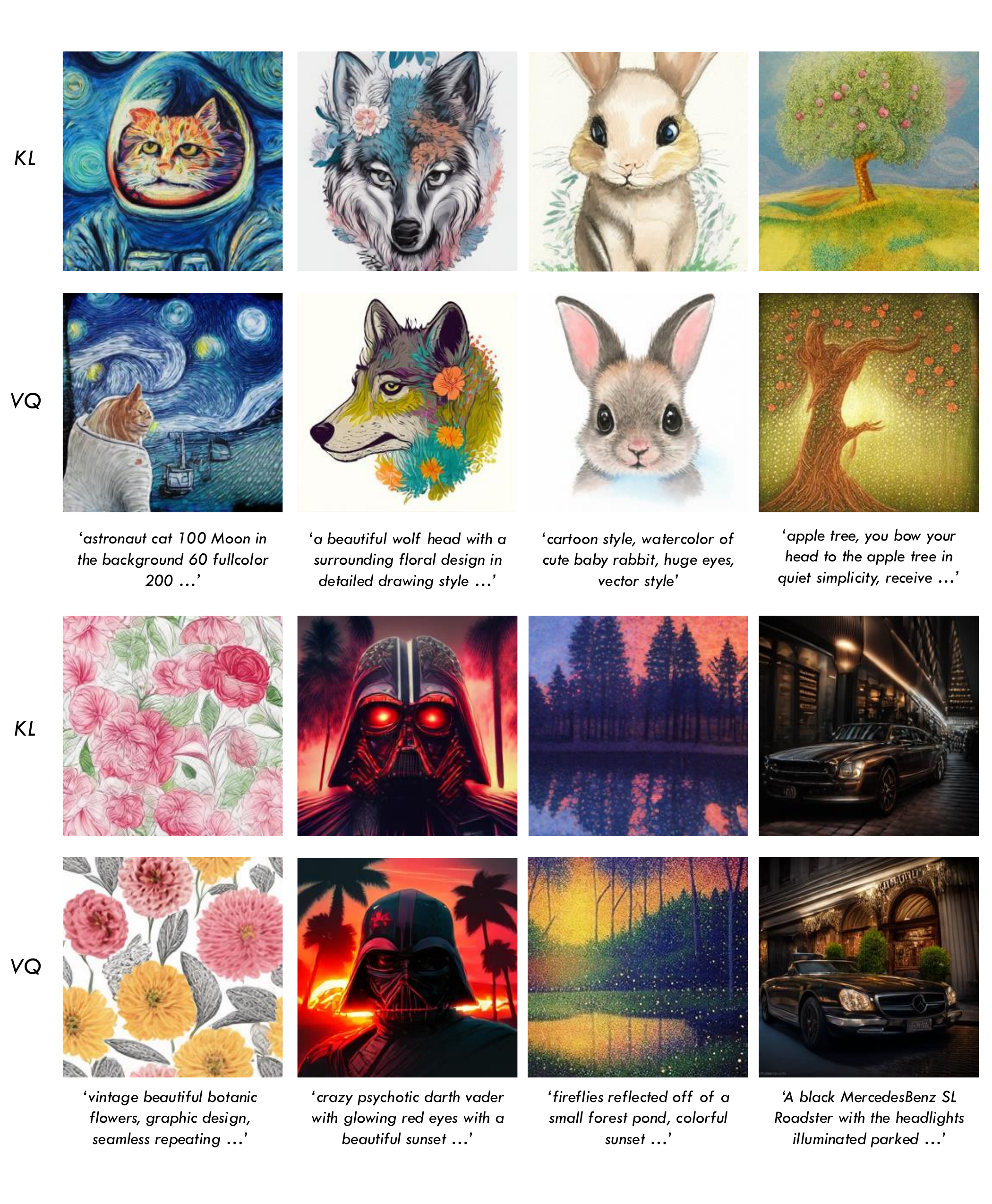}
    \caption{
    \textbf{Generated Images by \genmodelname with Different Tokenizer Types.}
For each caption, the top row displays images generated using continuous tokens (KL), while the bottom row shows images generated using discrete tokens (VQ). Long prompts are truncated for brevity.
    }
    \label{fig:vis_tok}
\end{figure*}

\noindent \textbf{Performance Comparisons Between VQ and KL Variants.}
The KL variant of \genmodelname consistently outperforms the VQ variant on MJHQ-30K FID but performs slightly worse on GenEval's overall score. We hypothesize that the KL variant excels in generating diverse, high-aesthetic images, contributing to improved FID on MJHQ-30K. However, it falls behind on GenEval, which emphasizes object-focused compositional properties such as position, count, and color. In contrast, the VQ variant, constrained by a finite codebook, generates less diverse but more compositionally accurate images, leading to higher scores on GenEval.
Fig.~\ref{fig:vis_tok} visually compares generated samples, where the KL variant demonstrates slightly better overall generation quality.

\noindent \textbf{Training and Inference Cost Comparisons Between VQ and KL Variants.}
The VQ variant of \genmodelname benefits from faster training and significantly faster inference, primarily due to inherent differences in the diffusion process used in the KL variant. While the KL variant excels in generating more diverse and higher-aesthetic images, this advantage comes with increased computational demands. 
To address this gap in training and inference efficiency, we employ 128 tokens for the VQ variant and 32 tokens for the KL variant, effectively controlling the training cost to remain at a comparable level, as shown in Tab.~4 of the main paper.

\input{sec/tables/t2i_coco_full}

\input{sec/tables/vae_imagenet_main}

\section{Zero-Shot Text-to-Image Generation Results on COCO}
\label{sec:coco}
In~\tabref{tab:t2i_coco}, we evaluate zero-shot text-to-image generation on the COCO dataset~\cite{lin2014microsoft} by randomly sampling 30K image-caption pairs from the COCO 2014 validation split and reporting the FID, as is standard in the literature. Since the fine-tuning stage of \genmodelname often generates more aesthetically appealing images that deviate from the COCO dataset distribution, we perform the evaluation using \genmodelname at the pre-training stage to ensure consistency with the dataset's characteristics. Notably, \genmodelname-L (KL variant with continuous tokens) achieves an FID-30K of 9.66, while \genmodelname-XL (KL variant) further improves to 8.98. These results demonstrate that \genmodelname achieves performance comparable to other state-of-the-art text-to-image models, highlighting its effectiveness even in the zero-shot setting.

\section{KL variant of TiTok on ImageNet}
\label{sec:mar_imagenet}
We evaluate the KL variant of \basemodelname as a drop-in replacement for standard 2D VAEs~\cite{kingma2013auto,li2024autoregressive} in class-conditional image generation on ImageNet~\cite{deng2009imagenet}. Results, reported in~\tabref{tab:imagenet_256}, are based on the MAR~\cite{li2024autoregressive} framework with its base model, after 400 epochs using unchanged MAR hyper-parameters. MAR with \basemodelname (our KL variant) achieves significant training time reductions (3.8$\times$ with 64 tokens, 2.7$\times$ with 128 tokens) and inference speedups (8.1$\times$ with 64 tokens, 3.2$\times$ with 128 tokens), thanks to its efficient 1D token design. Despite the substantial reduction in computational overhead, MAR with \basemodelname maintains performance comparable to MAR with conventional 2D VAEs using 256 tokens, highlighting \basemodelname's potential as an efficient and robust image tokenizer for class-conditional generation.

\begin{figure*}
    \centering
    \includegraphics[width=0.95\linewidth]{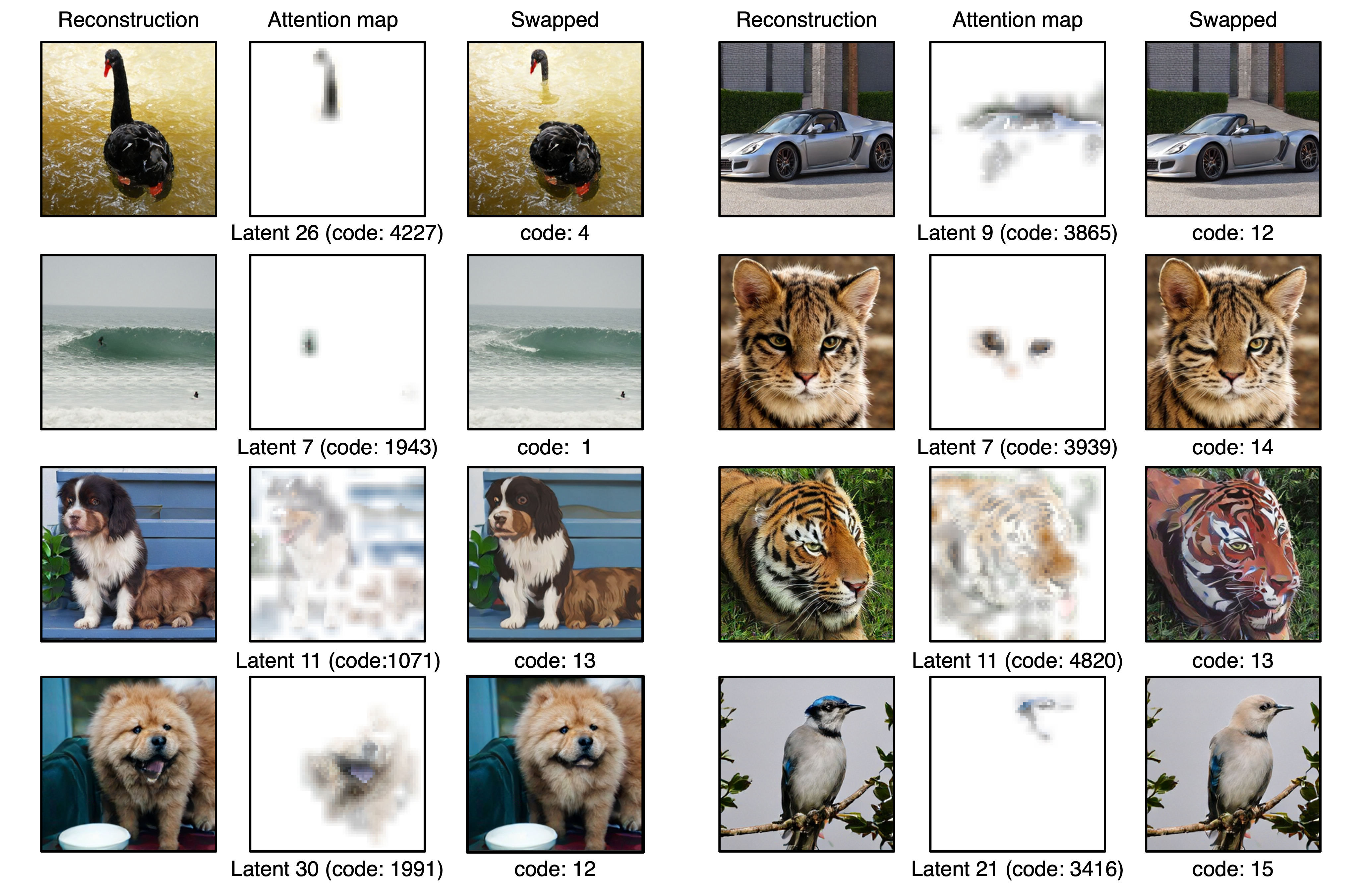}
    \vspace{-2mm}
    \caption{
    \textbf{Visualization of Latent Token Attention Map and Latent Code Swapping.} The results are from VQ variant of \modelname with 32 tokens. Each latent token attends to prominent semantic and swapping the code leads to appearance changes in the corresponding semantic entity that the latent token focuses on.
    }
    \vspace{-2mm}
    \label{fig:tokenswap_supp}
\end{figure*}
\section{Qualitative Examples of Latent Code Swapping}
\label{sec:more_swap_vis}
\figref{fig:tokenswap_supp} provides additional visualizations of latent code swapping in VQ variants of \modelname. The results demonstrate that our proposed 1D tokenization encodes images such that each token captures meaningful semantic elements, enabling image manipulation through latent token swapping without even requiring the generator.
This semantically rich tokenization explains why 1D tokenization achieves far more efficient encoding than its 2D counterpart while preserving competitive reconstruction fidelity: the encoder dynamically allocates tokens to perceptually informative regions in the image.

\section{Qualitative Examples of \genmodelname}
\label{sec:more_vis}
\begin{figure*}[]
    \centering
    \includegraphics[width=0.8\linewidth]{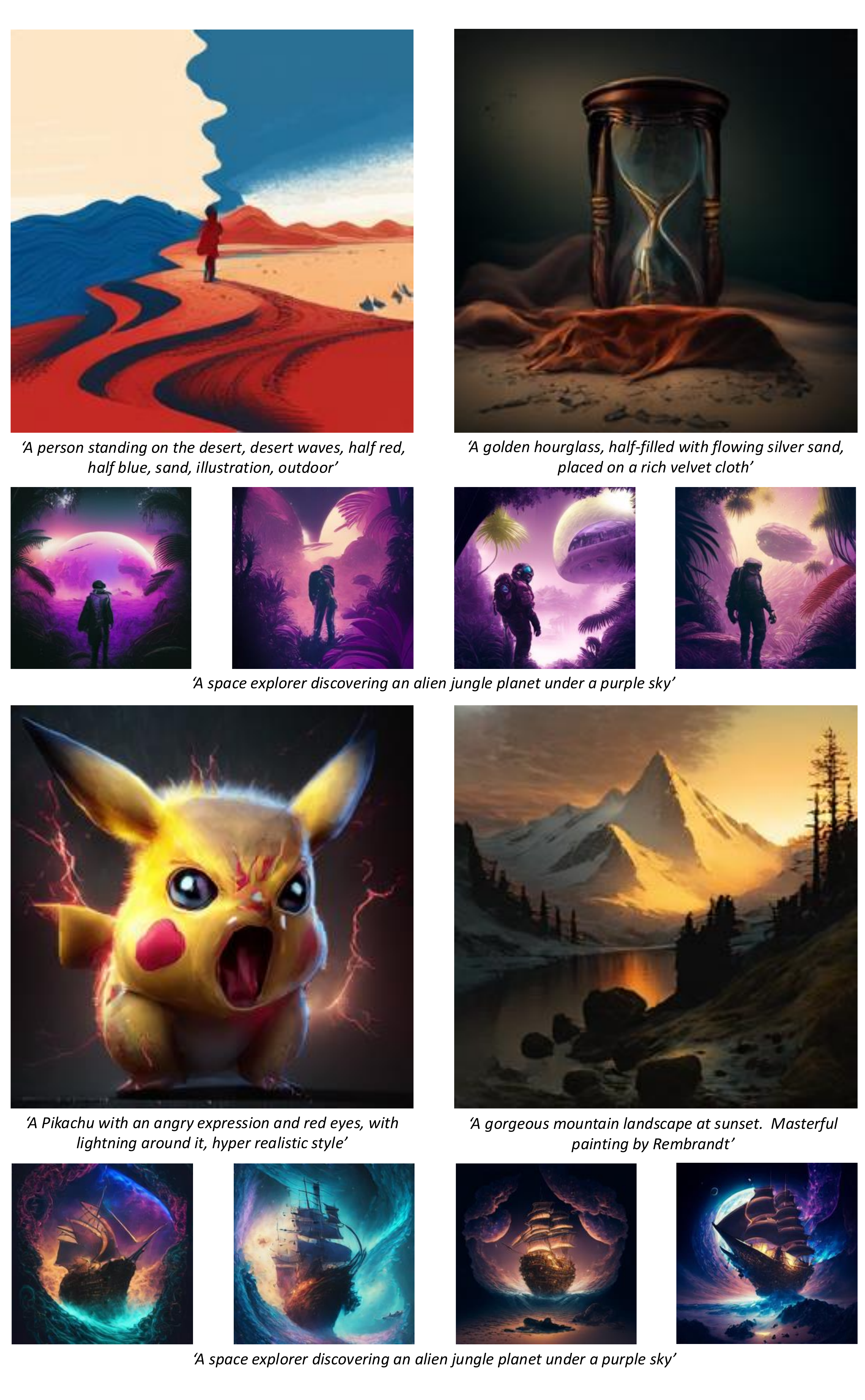}
    \vspace{-3ex}
    \caption{
    \textbf{Qualitative examples of Text-to-Image (T2I) Generation with \genmodelname.}
    \genmodelname, equipped with the efficient and compact text-aware 1D tokenizer \modelname, generates high-fidelity and diverse images.
    }
    \label{fig:vis_tok_1}
\end{figure*}

\begin{figure*}[]
    \centering
    \includegraphics[width=0.8\linewidth]{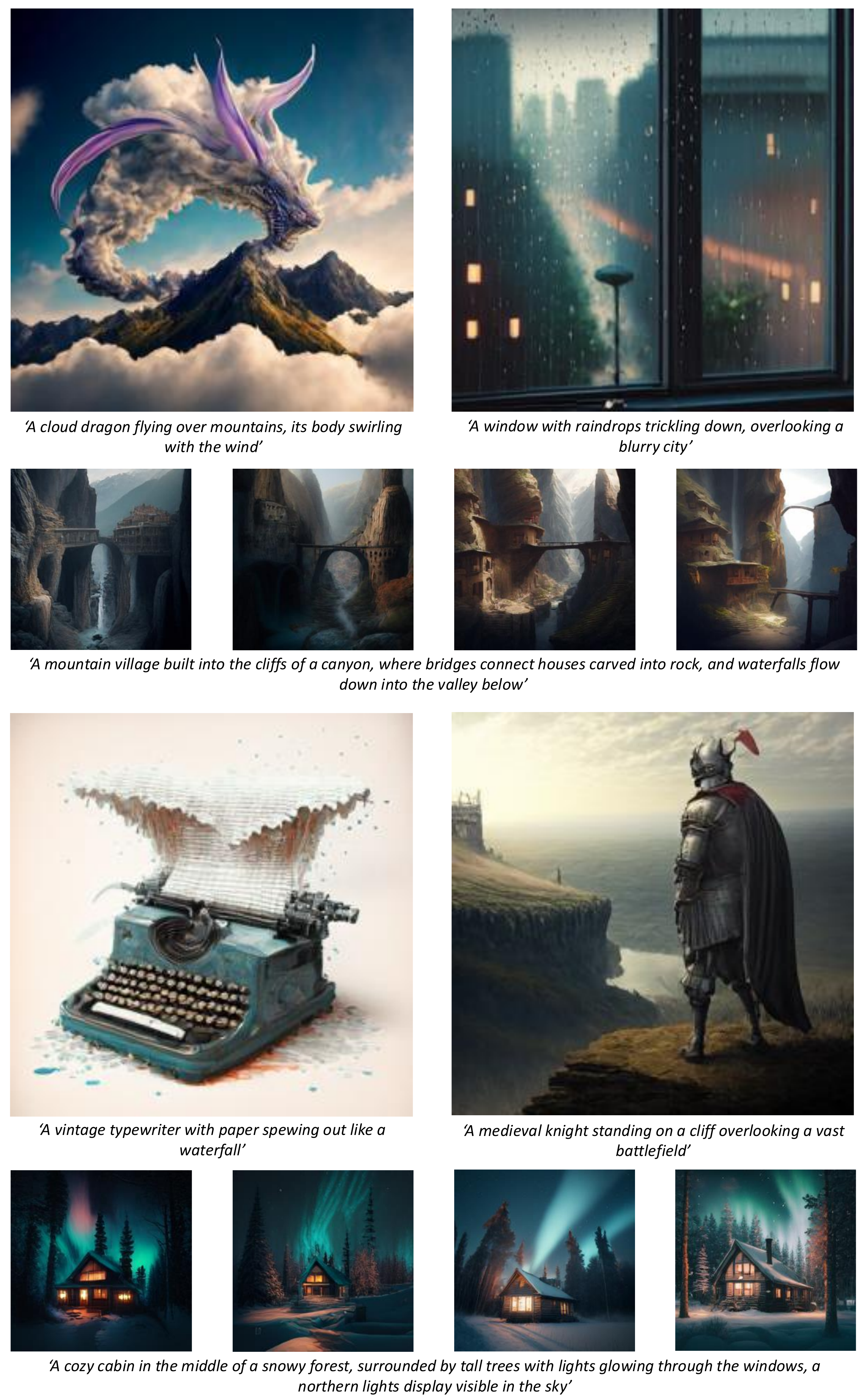}
    \vspace{-2.5ex}
    \caption{
    \textbf{Qualitative examples of Text-to-Image (T2I) Generation with \genmodelname.}
    \genmodelname, equipped with the efficient and compact text-aware 1D tokenizer \modelname, generates high-fidelity and diverse images.
    }
    \label{fig:vis_tok_2}
\end{figure*}

\begin{figure*}[]
    \centering
    \includegraphics[width=0.8\linewidth]{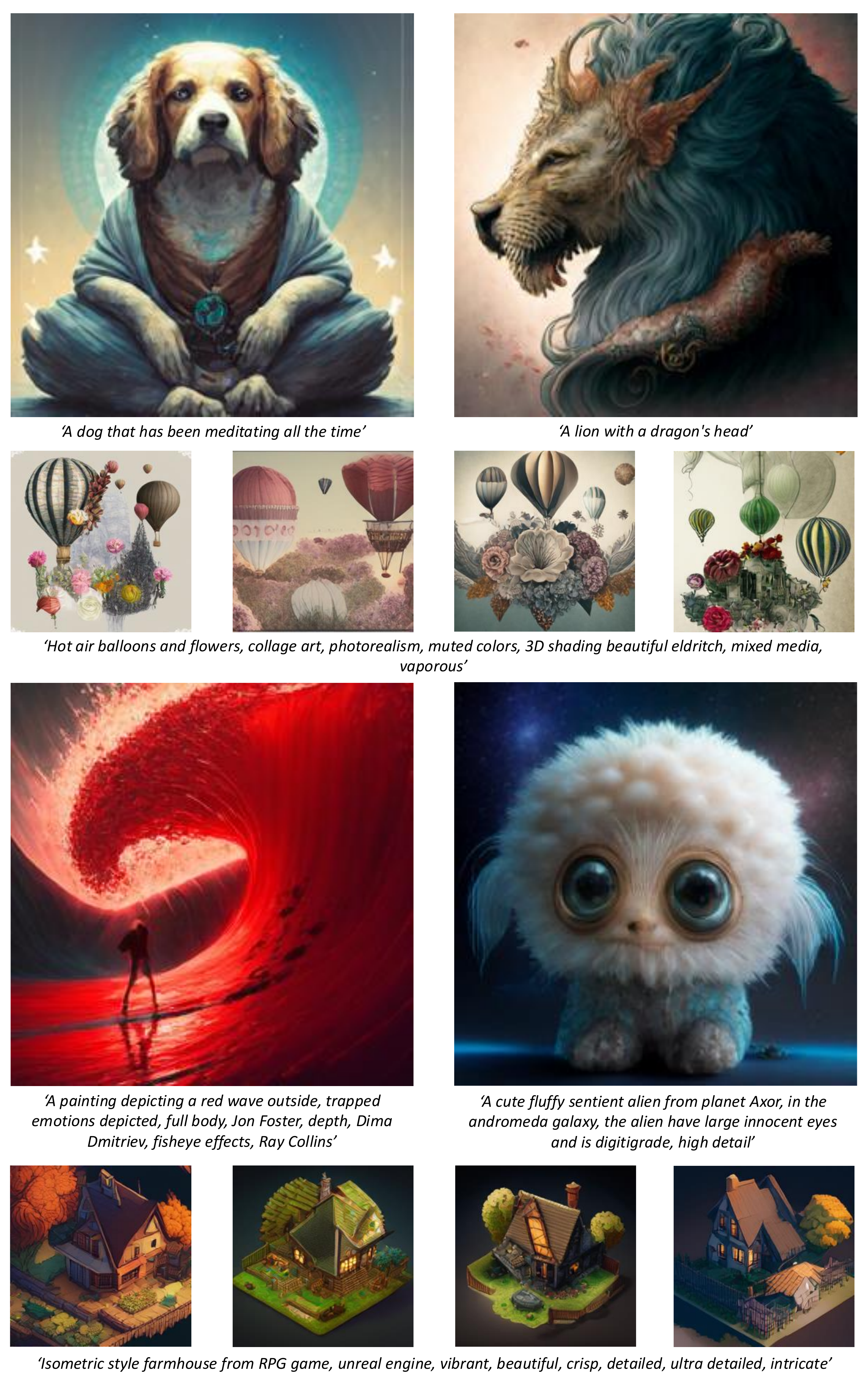}
    \vspace{-3ex}
    \caption{
    \textbf{Qualitative examples of Text-to-Image (T2I) Generation with \genmodelname.}
    \genmodelname, equipped with the efficient and compact text-aware 1D tokenizer \modelname, generates high-fidelity and diverse images.
    }
    \label{fig:vis_tok_3}
\end{figure*}

\begin{figure*}[]
    \centering
    \includegraphics[width=0.8\linewidth]{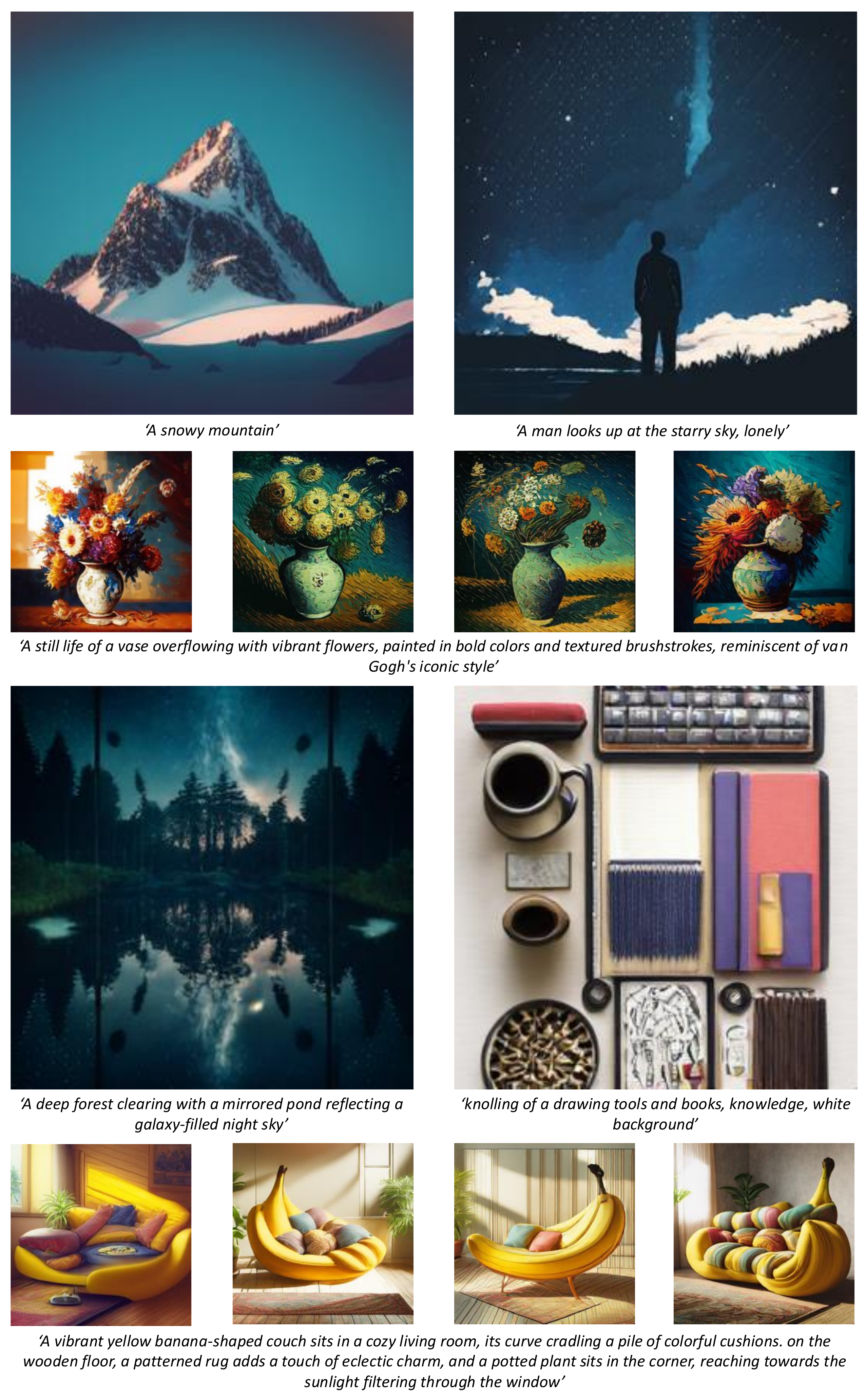}
    \vspace{-2.5ex}
    \caption{
    \textbf{Qualitative examples of Text-to-Image (T2I) Generation with \genmodelname.}
    \genmodelname, equipped with the efficient and compact text-aware 1D tokenizer \modelname, generates high-fidelity and diverse images.
    }
    \label{fig:vis_tok_4}
\end{figure*}

\begin{figure*}[]
    \centering
    \includegraphics[width=0.8\linewidth]{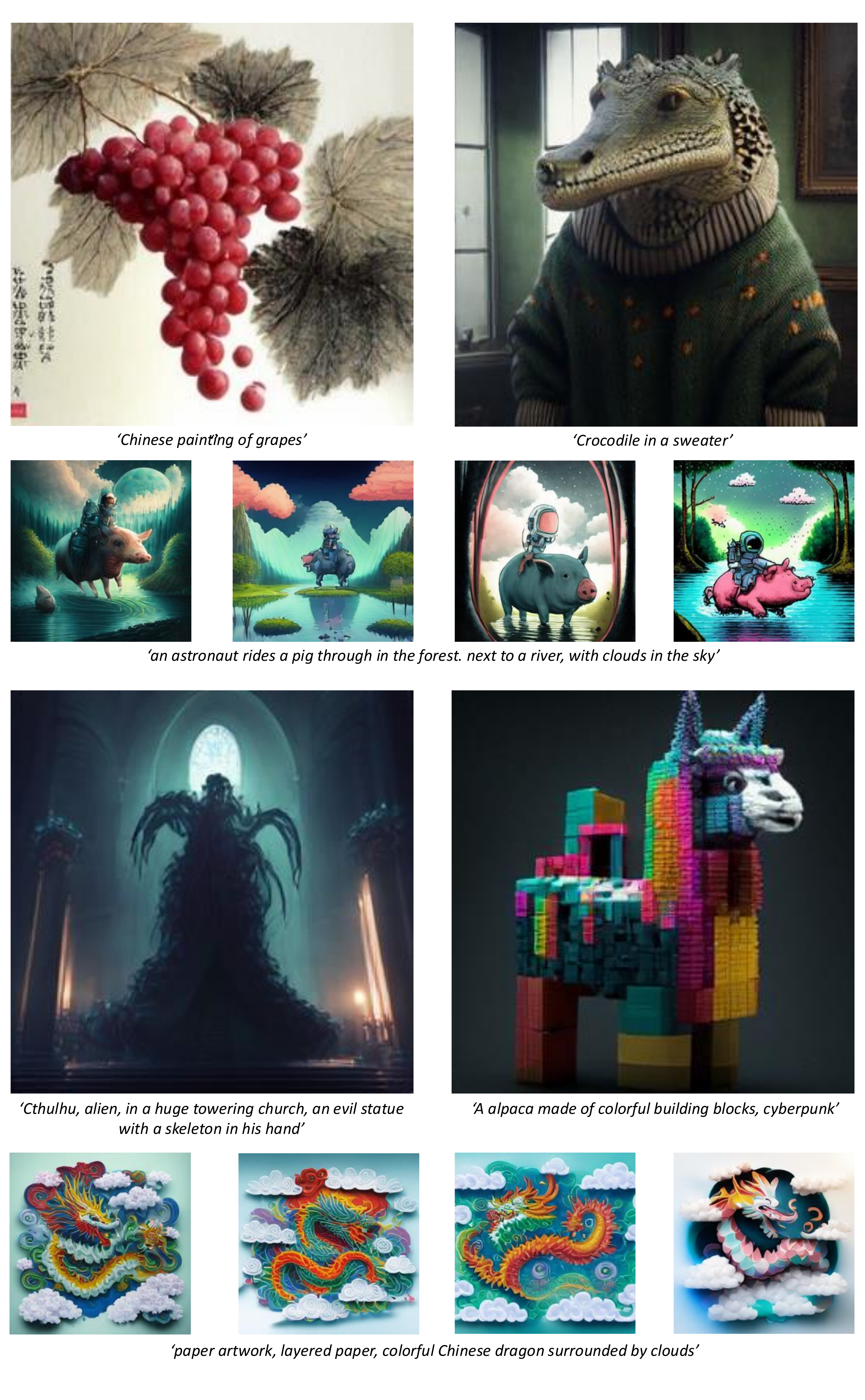}
    \vspace{-3ex}
    \caption{
    \textbf{Qualitative examples of Text-to-Image (T2I) Generation with \genmodelname.}
    \genmodelname, equipped with the efficient and compact text-aware 1D tokenizer \modelname, generates high-fidelity and diverse images.
    }
    \label{fig:vis_tok_5}
\end{figure*}

\begin{figure*}[]
    \centering
    \includegraphics[width=0.8\linewidth]{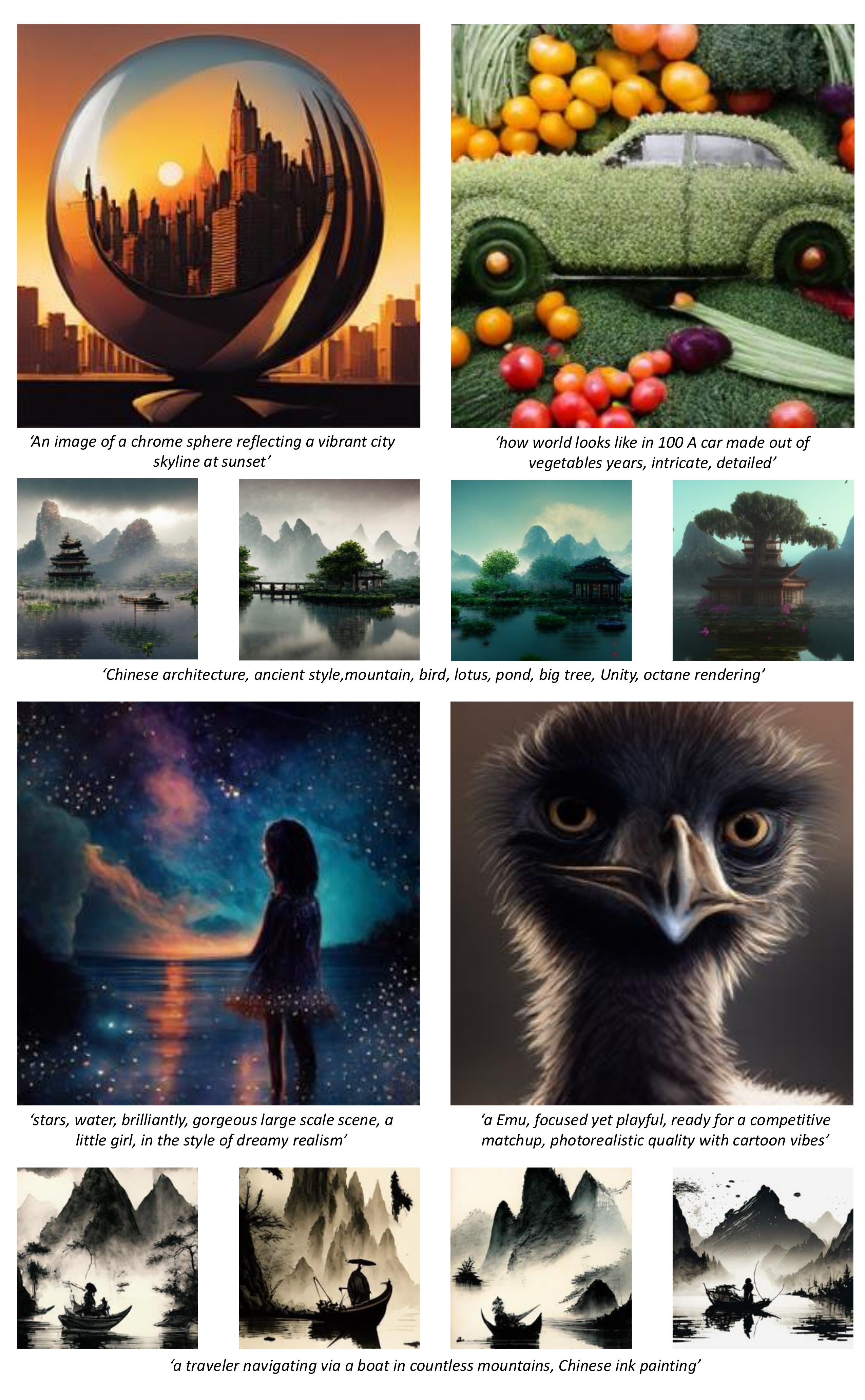}
    \vspace{-3ex}
    \caption{
    \textbf{Qualitative examples of Text-to-Image (T2I) Generation with \genmodelname.}
    \genmodelname, equipped with the efficient and compact text-aware 1D tokenizer \modelname, generates high-fidelity and diverse images.
    }
    \label{fig:vis_tok_6}
\end{figure*}

\begin{figure*}[]
    \centering
    \includegraphics[width=0.8\linewidth]{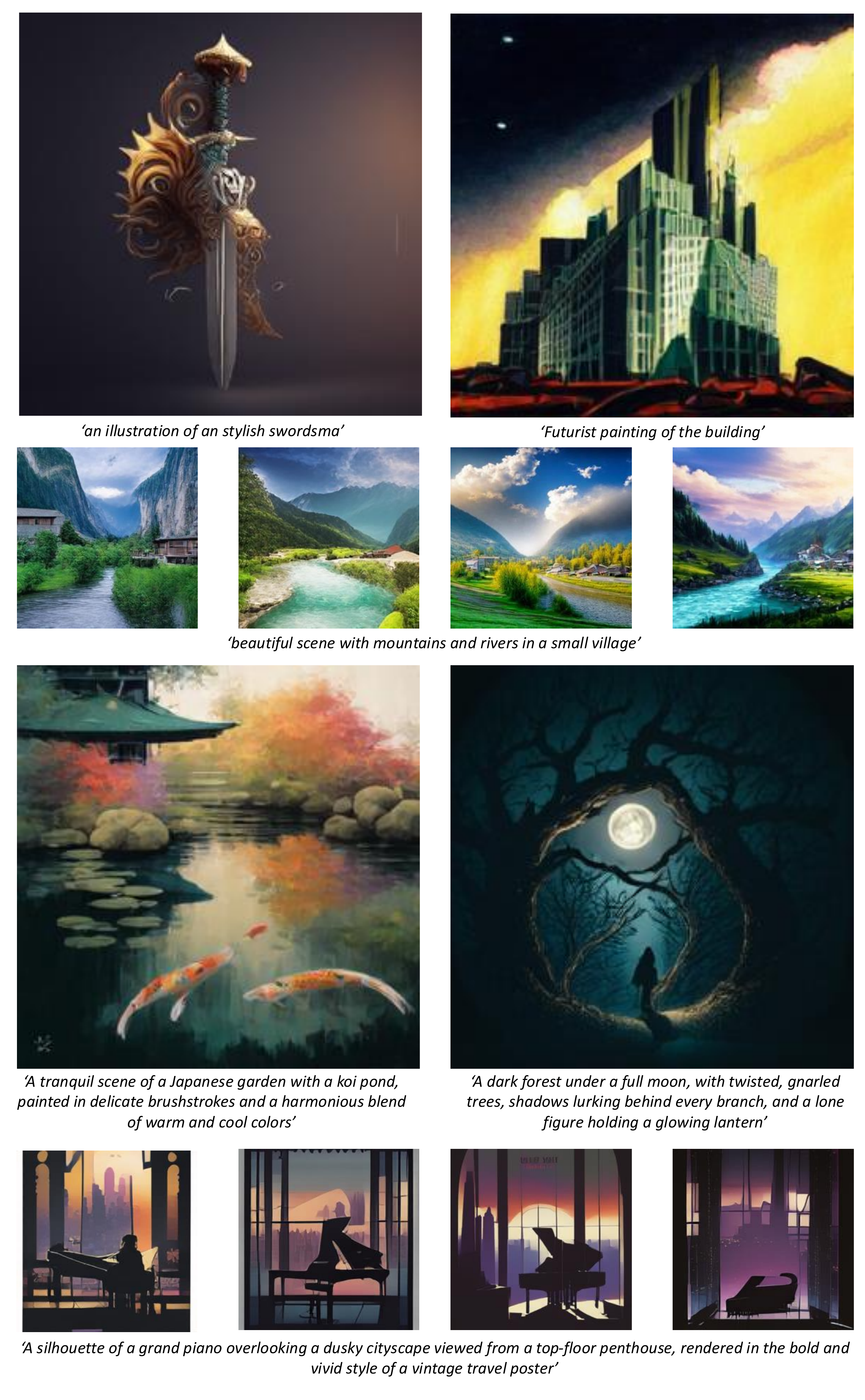}
    \vspace{-3ex}
    \caption{
    \textbf{Qualitative examples of Text-to-Image (T2I) Generation with \genmodelname.}
    \genmodelname, equipped with the efficient and compact text-aware 1D tokenizer \modelname, generates high-fidelity and diverse images.
    }
    \label{fig:vis_tok_7}
\end{figure*}

\figref{fig:vis_tok_1}, \figref{fig:vis_tok_2}, \figref{fig:vis_tok_3}, \figref{fig:vis_tok_4}, \figref{fig:vis_tok_5}, \figref{fig:vis_tok_6}, and~\figref{fig:vis_tok_7} showcase additional qualitative examples of text-to-image generation using \genmodelname. By utilizing the efficient and compact text-aware tokenizer \modelname, \genmodelname demonstrates its ability to produce high-fidelity and diverse images.

\section{Limitations and Future Work}
\label{sec:limitations}
While MaskGen achieves competitive generation quality and benchmark scores comparable to recent text-to-image models, including those leveraging proprietary training data, we acknowledge several aspects for future exploration.

First, the current KL variant of \genmodelname is designed to use 32 tokens. While increasing the token count improves tokenization quality, leading to better-reconstructed samples, it also significantly raises training costs due to longer convergence times. Additionally, scaling up the generator remains a challenge, as the current \genmodelname-XL is constrained to 1.1B parameters due to limited computational resources.

Second, the current implementation of \genmodelname operates at a resolution of $256\times256$. However, the scalability of its core architectural design—1-dimensional tokenization and masked generation—has been validated in high-resolution implementations like Muse~\cite{chang2023muse}.

This work emphasizes establishing a fully open-source, open-data text-to-image masked generative model using compact text-aware 1-dimensional tokenization. Future work will focus on optimizing convergence speed, model scaling up, and enabling high-resolution outputs.

\clearpage

\clearpage

%% file: sec/tables/dataset_detail.tex
\begin{table*}[!t]
\small
\centering
\tablestyle{2.0pt}{1.0}
\scalebox{1.2}{
\begin{tabular}{c|c|ccc|c|c}
\multirow{2}{*}{model} & \multirow{2}{*}{dataset} & \multicolumn{3}{c|}{filtering} & \multirow{2}{*}{recaptioning} & \multirow{2}{*}{samples} \\
 &  & resolution & aesthetic & watermark & & \\ \shline
\modelname: tokenizer& DataComp~\cite{gadre2024datacomp} & \cmark & & & & 685.8M \\ \hline

\multirow{3}{*}{\genmodelname: pre-training} & DataComp~\cite{gadre2024datacomp} & \cmark & \cmark~(5.0) & \cmark & & 219.8M\\ 
 & CC12M~\cite{changpinyo2021conceptual} & \cmark & \cmark~(5.0) & \cmark & & 4.8M \\ 
 & LAION-aesthetic~\cite{laion-aesthetic} & \cmark & & \cmark & & 28.3M\\ \hline
 
 \multirow{5}{*}{\genmodelname: fine-tuning} & DataComp~\cite{gadre2024datacomp} & \cmark & \cmark~(6.0) & \cmark & \cmark & 3.6M \\ 
 & LAION-art~\cite{laion-art} & \cmark & & \cmark & \cmark & 4.2M \\ 
 & LAION-pop~\cite{laion-pop} & \cmark & & \cmark & \cmark & 0.4M \\
 & DALLE3-1M~\cite{Egan_Dalle3_1_Million_2024} &  &  &  & & 1.0M \\
 & JourneyDB~\cite{sun2024journeydb} &  &  &  & & 4.1M \\ 
\end{tabular}
}
\vspace{-2mm}
\caption{
\textbf{Training Data Details.}
Filtering criteria applied to each publicly available dataset include resolution (aspect ratio < 2 and longer side $\ge$ 256), aesthetic score (predicted score exceeding the specified value in parentheses), and watermark detection (removal of images predicted to contain watermarks). For datasets with noisy web-crawled captions, Molmo~\cite{deitke2024molmo} is used for recaptioning.
The final column shows the number of text-image pairs remaining after filtering.
}
\vspace{-4mm}
\label{tab:dataset_detail}
\end{table*}

%% file: sec/tables/prompts.tex
\definecolor{backcolour}{rgb}{0.95,0.95,0.92}
\begin{figure}
\lstset{
basicstyle=\ttfamily\footnotesize,
breaklines=true,                %
breakatwhitespace=true,         %
postbreak=\mbox{\space},        %
prebreak=\mbox{\space},              %
breakindent=0pt,               %
columns=flexible,               %
showstringspaces=false,         %
xleftmargin=15pt,              %
captionpos=b,                  %
commentstyle=\color{gray},      %
upquote=true
}
\begin{lstlisting}
1. Describe the image in detail while considering the provided caption: '{original_caption}'. Correct any errors and improve the caption, ensuring the final description is in English and within 77 tokens. Return only the corrected caption.

2. Analyze the image and the caption '{original_caption}'. Write a detailed and accurate description of the image in English, correcting any mistakes or low-quality aspects of the original caption. Keep the final caption under 77 tokens, and return only the caption.

3. Use the caption '{original_caption}' as a reference to create a detailed and improved description of the image in English. Correct any errors and make sure the new caption is concise and within 77 tokens. Return only the revised caption.

4. Given the image and the original caption '{original_caption}', describe the image in a detailed and accurate way in English, improving upon the original caption where necessary. Ensure the description is within 77 tokens. Return only the corrected caption.
\end{lstlisting}
\vspace{-2mm}
\caption{
    \textbf{Prompts Used for Recaptioning.} 
    One of four prompts is used to recaption each image, where \texttt{\{original\_caption\}} is replaced with the original image caption.}
\vspace{-2mm}
\label{fig:prompts}
\vspace{-3ex}
\end{figure}

%% file: sec/tables/training_hparams.tex
\newcommand*{\mytab}{\hspace*{0.35cm}}

\begin{table}
\small
\centering
\tablestyle{2.0pt}{1.0}
\scalebox{1.2}{
\begin{tabular}{l|cc}
 hyper-parmeters & discrete & continuous \\ \shline
 
optimizer & AdamW & AdamW \\
$\beta_1$ & 0.9 & 0.9 \\
$\beta_2$ & 0.96 & 0.95 \\
weight decay & 0.03 & 0.02 \\ 
 
 lr (pre-training) & 0.0004 & 0.0002 \\
 lr (fine-tuning) & 0.0001 & 0.0002 \\
 lr scheduling & cosine & constant\\
 lr warmup steps & 10K & 50k \\ 

 batch size & 4096 & 2048 \\
 training steps (pre-training) & 500K  & 1000k \\
 training steps (fine-tuning) & 250K  & 500k \\
 
\end{tabular}
}
\vspace{-2mm}
\caption{
\textbf{Training Hyper-parameters for  \genmodelname.}
}
\vspace{-2mm}
\label{tab:training_hparams}
\end{table}

%% file: sec/tables/ablation_text-type.tex
\begin{table}
\centering
\small
\tablestyle{2.0pt}{1.0}
\scalebox{1.1}{
\begin{tabular}{lcccccc}
tokenizer & arch & \#tokens & text guidance & rFID$\downarrow$ & IS$\uparrow$ \\
\shline
\multirow{2}{*}{\modelname} & \multirow{2}{*}{KL} & \multirow{2}{*}{32} & ID & 1.62 & 213.6 \\
 & & & Embedding & 1.53 & 222.0 \\
\end{tabular}
}
\vspace{-2mm}
\caption{\textbf{Ablation on Text Guidance Type.}
Models are trained on DataComp and zero-shot evaulated on ImageNet validation set.
ID refers to numerical IDs extracted by CLIP text tokenizer, Embedding denotes text features extracted by CLIP text encoder.}
\vspace{-2mm}
\label{tab:ablation_text-type}
\end{table}

%% file: sec/tables/ablation_text-place.tex
\begin{table}
\centering
\small
\tablestyle{2.0pt}{1.0}
\scalebox{1.1}{
\begin{tabular}{ccc|cc|cc}
\multirow{2}{*}{arch} & \multicolumn{2}{c|}{tokens} & \multicolumn{2}{c|}{Encoder + Decoder} & \multicolumn{2}{c}{Decoder Only} \\
 & \# & c & rFID$\downarrow$ & IS$\uparrow$ & rFID$\downarrow$ & IS$\uparrow$ \\ \hline
\multirow{3}{*}{KL} & 32 & 16 & 1.65 & 218.4 & 1.53 & 222.0 \\
& 64 & 16 & 1.39 & 221.5 & 1.47 & 220.7 \\
& 128 & 16 & 0.92 & 227.1 & 0.90 & 227.7 
\end{tabular}
}
\vspace{-2mm}
\caption{\textbf{Ablation on Text Guidance Place.}
In the \modelname design, we ablate on adding the text guidance to both encoder and decoder or just decoder.
Adding text guidance to only the decoder results in similar reconstruction performances but enjoys a simpler structure.
Models are trained on DataComp and zero-shot evaulated on ImageNet validation set.}
\vspace{-2mm}
\label{tab:ablation_text-place}
\end{table}

%% file: sec/tables/ablation_token_count.tex
\begin{table}
\small
\centering
\tablestyle{2.0pt}{1.0}
\scalebox{1.2}{
\begin{tabular}{cc|c|cc|c|c}
     &           &         &     &  & MJHQ-30K & GenEval \\
arch & generator& \#tokens & T$\downarrow$ & I$\uparrow$ & FID$\downarrow$ & Overall$\uparrow$ \\ \shline
\multirow{3}{*}{VQ} & \multirow{3}{*}{\genmodelname-L}   & 32 & 16.0 & 47.6 & 9.11 & 0.43 \\
&  & 64 & 17.5 & 40.2 & 7.85 & 0.50 \\
&  & 128 & 20.0 & 30.3 & 7.74 & 0.53 \\
\end{tabular}
}
\vspace{-2mm}
\caption{
\textbf{Zero-Shot Text-to-Image Generation Results on MJHQ-30K and GenEval with Varying Number of Tokens.}
\genmodelname achieves better generation quality with more tokens but incurs longer training times and slower inference speeds.
T: Generator training cost, measured in 8 A100 days using float16 precision. I: Generator inference throughput, measured in samples per second on a single A100 with batch size 64 using float16 precision.
}
\vspace{-4mm}
\label{tab:ablation_token_count}
\end{table}

%% file: sec/tables/t2i_coco_full.tex
\begin{table}[!t]
\small
\centering
\tablestyle{2.0pt}{1.0}
\scalebox{0.8}{
\begin{tabular}{lc|lc|c|c}
tokenizer & arch & generator & \#params & open-data & FID-30K$\downarrow$ \\ \hline
MAGVIT-v2~\cite{xie2024show} & VQ & Show-o~\cite{xie2024show} & 1.3B & \cmark & 9.24 \\
\hline
\modelname & VQ & \genmodelname-L (ours) & 568M & \cmark & 13.62 \\
\modelname & VQ & \genmodelname-XL (ours) & 1.1B & \cmark & 13.01 \\
\hline
\hline
VAE~\cite{rombach2022high} & KL & LDM~\cite{rombach2022high} & 1.4B & \cmark & 12.64 \\
VAE~\cite{rombach2022high} & KL & Stable-Diffusion-1.5~\cite{rombach2022high} & 860M & \cmark & 9.62 \\
VAE~\cite{rombach2022high} & KL & PixArt-$\alpha$~\cite{chen2024pixartalpha} & 630M & \xmark & 7.32 \\
\hline
\modelname & KL & \genmodelname-L (ours) & 568M + 44M & \cmark & 9.66 \\
\modelname & KL & \genmodelname-XL (ours) & 1.1B + 69M & \cmark & 8.98 \\
\end{tabular}
}
\vspace{-2mm}
\caption{
\textbf{Zero-Shot Text-to-Image Generation Results on COCO-30K.}
Comparison of \genmodelname with state-of-the-art \textit{open-weight} models.
}
\vspace{-2mm}
\label{tab:t2i_coco}
\end{table}

%% file: sec/tables/vae_imagenet_main.tex
\begin{table}
\centering
\small
\tablestyle{2.0pt}{1.0}
\scalebox{1.0}{
\begin{tabular}{lcccc|lcccc}
\multirow{2}{*}{tokenizer} & \multirow{2}{*}{arch} & \multicolumn{2}{c}{tokens} & \multirow{2}{*}{rFID$\downarrow$} & \multirow{2}{*}{generator} & \multirow{2}{*}{gFID$\downarrow$} & \multirow{2}{*}{IS$\uparrow$} & \multirow{2}{*}{T$\downarrow$} & \multirow{2}{*}{I$\uparrow$} \\
 & & \# & c &  &  &  &  &  &  \\
\shline
VAE~\cite{li2024autoregressive} & KL & 256 & 16 & 0.54 & MAR~\cite{li2024autoregressive} & 2.45 & 275.5 & 8.0 & 1.0 \\
\hline
\multirow{2}{*}{\basemodelname (ours)} & \multirow{2}{*}{KL} & 64 & 16 & 1.54 & \multirow{2}{*}{MAR~\cite{li2024autoregressive}} & 2.96 & 246.9 & 2.1 & 8.1 \\
 & & 128 & 16 & 1.31 &  & 2.70 & 252.9 & 3.2 & 3.2 \\
\end{tabular}
}
\caption{\textbf{Class-conditional ImageNet-1K $256\times 256$ Generation Results Evaluated with ADM~\cite{dhariwal2021diffusion}, using continuous tokens (\ie, KL architecture).}
\#: Number of tokens. c: Channels of continuous tokens. T: Generator training cost, measured in 8 A100 days using float32 precision. I: Generator inference throughput, measured in samples per second on a single A100 with float32 precision.
}
\label{tab:imagenet_256}
\end{table}